\pdfoutput=1
\def\ACLPreprintMode{1}
\documentclass[11pt]{article}

\ifdefined\ACLFinalMode
  \usepackage[final]{acl}
\else\ifdefined\ACLPreprintMode
  \usepackage[preprint]{acl}
\else
  \usepackage[review]{acl}
\fi\fi

\usepackage{times}
\usepackage{latexsym}
\usepackage[T1]{fontenc}
\usepackage[utf8]{inputenc}
\usepackage{microtype}
\usepackage{inconsolata}
\usepackage{graphicx}
\usepackage{subcaption}
\usepackage{booktabs}
\usepackage{tabularx}
\usepackage{enumitem}
\usepackage{array}
\usepackage{multirow}
\usepackage{float}
\usepackage{placeins}  
\usepackage{amsmath}
\usepackage{amssymb}

\definecolor{deepseekblue}{HTML}{24607A}
\definecolor{lunagreen}{HTML}{3B7F5C}
\definecolor{glmcharcoal}{HTML}{333333}
\newcommand{\deepseekname}[1]{\textcolor{deepseekblue}{\textbf{#1}}}
\newcommand{\lunaname}[1]{\textcolor{lunagreen}{\textbf{#1}}}
\newcommand{\glmname}[1]{\textcolor{glmcharcoal}{\textbf{#1}}}

\newcolumntype{Y}{>{\raggedright\arraybackslash}X}
\definecolor{chipfill}{HTML}{E4EAED}
\newcommand{\mainlvl}[1]{{\setlength{\fboxsep}{1.4pt}\colorbox{chipfill}{\strut\textbf{#1}}}}
\newcommand{\sep}{\unskip\nobreak\enspace{\color{gray}\textbar}\hspace{0.5em plus 0.3em}\ignorespaces}
\newcommand{\lapse}{\textsc{Lapse}}

\ifdefined\ACLFinalMode
  \newcommand{\lapseurl}{\url{https://github.com/aimsresearchlab/lapse}}
\else\ifdefined\ACLPreprintMode
  \newcommand{\lapseurl}{\url{https://github.com/aimsresearchlab/lapse}}
\else
  \newcommand{\lapseurl}{\url{https://anonymous.4open.science/r/lapse}}
\fi\fi
\ifdefined\ACLFinalMode\def\lapsenamed{}\fi
\ifdefined\ACLPreprintMode\def\lapsenamed{}\fi
\ifdefined\lapsenamed
  \newcommand{\lapselinks}{\par\smallskip{\centering\small
    \linkicon{github}{github.com/aimsresearchlab/lapse}\\
    \linkicon{huggingface-color}{huggingface.co/datasets/vein05/lapse}\par}}
\else
  \newcommand{\lapselinks}{Code and data: \lapseurl.}
\fi

\graphicspath{{figures/}{figures/icons/}}
\newcommand{\mico}[1]{\raisebox{-0.2ex}{\includegraphics[height=1.6ex]{#1.pdf}}\,}

\title{Language Model Memory Flattens the Temporal Form of User Facts}

\author{Anonymous Authors}

\ifdefined\ACLPreprintMode
  \usepackage{aimsheader}

  \hypersetup{
    pdftitle={Memory Consolidation Flattens the Temporal Shape of User Facts},
    pdfauthor={Sugam Panthi, Muhaiminul Yeamin, Siyan Luo, Rabab Abdelfattah}
  }
\fi

\begin{document}
\ifdefined\ACLPreprintMode
  \twocolumn[{%
    \aimslinetitle{
      title   = {Memory Consolidation Flattens the Temporal Shape of User Facts},
      authors = {Sugam Panthi, Muhaiminul Yeamin, Siyan Luo, Rabab Abdelfattah},
      affil   = {AIMS Lab, The University of Southern Mississippi\\
        \texttt{\{sugam.panthi, muhaiminul.yeamin, siyan.luo, rabab.abdelfattah\}@usm.edu}},
      right   = {Preprint, version 2 \textbullet\ November 11, 2026},
    }
    \vspace{4pt}
  }]
\else
  \maketitle
\fi

\begin{abstract}
LLM agents with long-term memory store conversations as short notes that later models
read. A note can turn progressive ``I am working at Corvida'' into
simple-present ``The user works at Corvida,'' dropping the cue that the job
may be temporary and that the stored fact may go stale. We call this \emph{aspectual flattening} and measure it with
\lapse{}, a benchmark of matched user statements that differ only in temporal
form. Three writer models flattened the progressive statement but kept its
simple-present match in 210 of 381 pairs, never the reverse. This one-way loss
held in all 11 models tested and in the agent memory systems mem0, Graphiti, and
Letta. In exploratory tests, reader models judged a fact less likely to still
hold when only its stored verb was progressive. In a planned test,
the reader \mbox{GLM-5.2} acted without asking the user on 64\% of ``works''
notes but 40\% of ``is working'' notes. Gemini 3 Flash showed a smaller gap,
one reader added later showed the reverse, and others showed none. In a further exploratory test, free-text notes
erased this difference and notes that kept the user's words restored it. Agent memory systems should keep the
user's wording as evidence for later readers. \lapselinks
\end{abstract}

\section{Introduction}

Consider a user who says, ``I am working at Corvida.'' A memory writer, the
model or pipeline step that turns a conversation into stored notes, may store
\emph{The user works at Corvida}. The person and employer survive, but the cue
that the job may be temporary does not. The model that later reads the note
never sees the conversation, so the note is the only message it receives.
Months later, that reader has less evidence for deciding whether to check the
fact first. We call this transformation \emph{aspectual flattening}.

Memory systems already address storage, retrieval, decay, contradictions, and
timing
\citep{packer-etal-2023-memgpt,wu-etal-2024-longmemeval,chhikara-etal-2025-mem0}.
A timestamp records when the user spoke, but not whether the user presented
the job as temporary. A writer can keep the timestamp while erasing that
distinction.

Aspect is how a sentence presents the internal temporal structure of a
situation, for example as ongoing or as a standing state
\citep{vendler1957verbs,comrie1976aspect,friedrich-etal-2023-kind}.
\citet{moens-steedman-1988-temporal} argued that aspect can change the
temporal category of a proposition, and that any usable temporal database
queried in natural language must represent such categories. A long-term memory
store is one such database, and its notes are messages from one model to
another. We therefore treat memory writing as communication between models and
ask whether the note keeps the aspectual evidence that its reader would need.

Testing this question requires a comparison that changes the wording without
changing the situation. We build \lapse{} (\textbf{L}inguistic \textbf{A}spect
\textbf{P}ersistence and \textbf{S}tability \textbf{E}valuation), a controlled
benchmark of matched user statements (Table~\ref{tab:design}). Within a pair,
the surrounding words, situation, and elapsed time stay fixed; only the
temporal form changes. Controls test two other explanations for an apparent
effect: a model's general tendency to act on any stored fact, and content that
sounds temporary on its own. We fixed the main test and its analysis before
collecting data. After a human check of the labels, we stopped counting notes
that keep the progressive but add ``currently'' as a loss; the test passes
either way (Section~\ref{sec:scoring}).

\begin{figure*}[t]
  \centering
  \includegraphics[width=\textwidth]{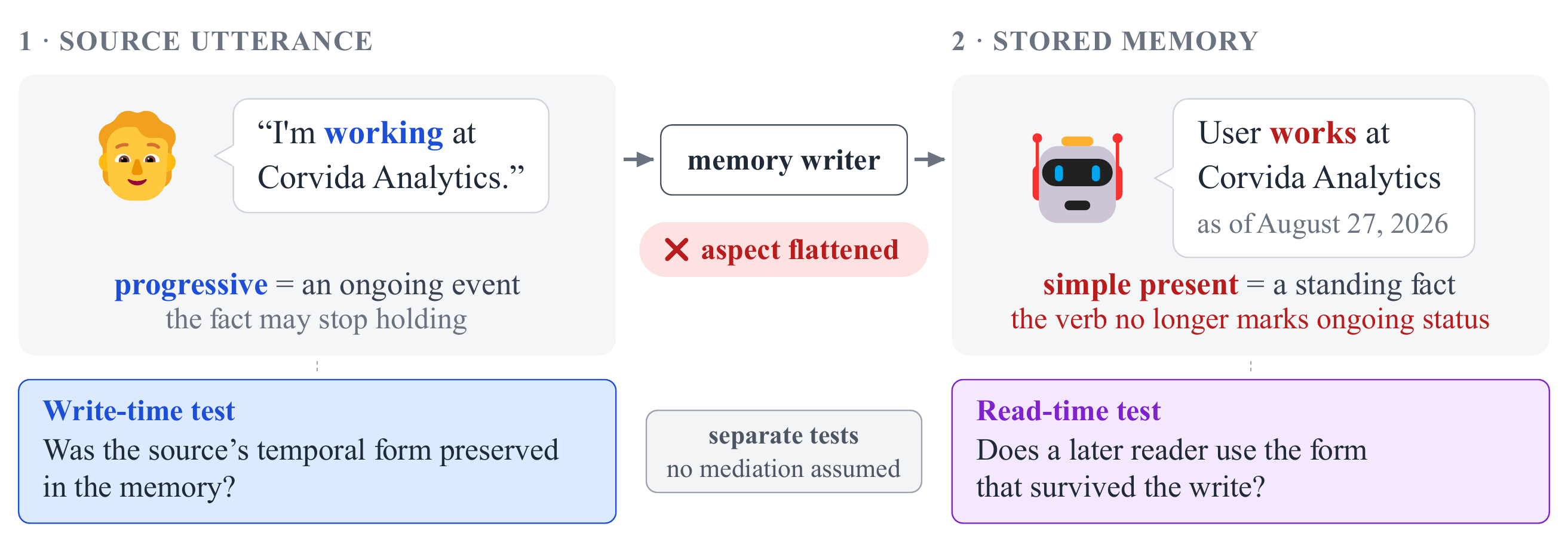}
  \caption{\lapse{} separates what a writer stores from how a reader uses it.
  Panels 1--2 show a note written by the released memory system mem0: the writer
  replaces progressive \emph{working} with simple-present \emph{works} but
  keeps a date.}
  \label{fig:overview}
\end{figure*}

Flattening is a miscommunication in which nothing false is said. ``The user
works at Corvida'' is true when written, but it no longer shows how the user
presented the job, and the reader cannot ask the writer what was meant. A
miscommunication can arise on the sender side or the receiver side, so we ask
two separate questions (Figure~\ref{fig:overview}): does the writer keep the
source cue, and does the reader use the cue that survives? Scoring the reader's
final behavior alone could not show which side lost the cue. The form of the
message is also a design choice, so we compare a free-text note with a
structured record that keeps the user's words (Section~\ref{sec:record}).

Our contributions are:
\begin{itemize}[leftmargin=*,itemsep=2pt,topsep=3pt]
  \item \textbf{Construct and benchmark.} We name \emph{aspectual flattening}
  and release \lapse{}, a benchmark of matched user statements that differ only
  in temporal form, with its stimulus builder, model outputs, labels, and
  analysis programs (\lapseurl).

  \item \textbf{Write-time result.} Memory writers flatten the progressive
  while keeping the matched simple present in 210 of 381 pairs, and never the
  reverse. The pattern holds in all 11 models tested and in the released
  memory systems mem0, Graphiti, and Letta.

  \item \textbf{Read-time evidence.} In exploratory tests, changing only the
  stored verb to the progressive lowers readers' estimates that a fact still
  holds. In a planned test, GLM-5.2 and Gemini 3 Flash then act without asking
  less often, and one reader added later does the reverse. In an exploratory
  test, free-text notes erased this difference and records that kept the
  user's words restored it.
\end{itemize}

\section{Related Work}

\begin{table*}[t]
\centering
\small
\renewcommand{\arraystretch}{1.25}
\begin{tabularx}{\textwidth}{@{}l c >{\raggedright\arraybackslash\hyphenpenalty=10000\exhyphenpenalty=10000}X >{\raggedright\arraybackslash\hyphenpenalty=10000}p{4.1cm}@{}}
\toprule
Factor & Levels & Values (\mainlvl{main test} highlighted) & Varied to test \\
\midrule
Form & 5 & \mainlvl{progressive}\sep\mainlvl{simple~present}\sep bounded\sep
perfect~progressive\sep perfect~simple & the manipulated variable \\
Frame & 8 & \mainlvl{all eight:} lodging\sep workplace\sep vehicle\sep class\sep
household\sep equipment\sep affiliate~role\sep project & whether it holds
across kinds of fact \\
Carrier & 6 & \mainlvl{life update}\sep after~a~request\sep bare~statement\sep
mid-message\sep formal~note\sep list~item & whether the surrounding
conversation matters \\
Gap & 6 & fresh~(2~days)\sep near~(42~days)\sep boundary\sep expired-soon\sep
\mainlvl{stale~(245~days)}\sep stale-long & fresh and boundary serve as
positive controls \\
Task & 7 & \emph{write:} \mainlvl{memory write}\sep guided~memory~write\newline
\emph{use:} direct~use\sep memory~use\sep guided~memory~use\newline \emph{ask:}
validity~question\sep send-or-check~choice & what is stored versus how it is
used \\
\midrule
\multicolumn{4}{@{}p{\textwidth}@{}}{\emph{Main test:} progressive vs.\ simple
present in all eight frames, with the life-update carrier, the stale gap, and
the memory-write task: 128 matched pairs per configuration.} \\
\bottomrule
\end{tabularx}
\caption{The \lapse{} design. An example pair is ``I am working at Corvida''
vs.\ ``I work at Corvida''; both members use the same verb. Appendix~\ref{app:materials}
gives every template.}
\label{tab:design}
\end{table*}

\paragraph{Long-term agent memory.}
Generative Agents, MemoryBank, MemGPT, RecallM, THEANINE, A-MEM, Mem0, and
Graphiti store and retrieve long-lived information in different forms
\citep{park-etal-2023-generative,zhong-etal-2024-memorybank,
packer-etal-2023-memgpt,kynoch-etal-2023-recallm,
ong-etal-2025-theanine,xu-etal-2025-amem,chhikara-etal-2025-mem0,
rasmussen-etal-2025-zep}. Their temporal mechanisms include timestamps,
recency, decay, graph chronology, and consolidation. Recent systems also build
semantic timelines or attach lifecycle relations and policies to extracted facts
\citep{su-etal-2026-temporal,yacoubi-etal-2026-memlace,
mullick-tuzun-2026-fortunate}. These methods reason over stored events,
metadata, or relations among memories. \lapse{} asks an earlier question:
does memory writing preserve temporal evidence already present in the user's
words?

\paragraph{Long-context and temporal evaluation.}
LoCoMo and LongMemEval test long-horizon recall, temporal reasoning, updates,
and abstention \citep{maharana-etal-2024-locomo,wu-etal-2024-longmemeval}.
Their user turns include temporary-form statements, but we found no question
that requires noticing that form (Appendix~\ref{app:ecology}).
Which stored form receives retrieval credit can also change conclusions drawn
from these benchmarks \citep{panthi-abdelfattah-2026-ranking}.
TVCP predicts validity from historical text and timestamps
\citep{wenzel-jatowt-2024-tvcp}, while Chronocept evaluates temporal-concept
understanding \citep{goel-etal-2026-chronocept}. Most closely, STALE studies
stale personalized knowledge after a later observation changes the world
\citep{chao-etal-2026-stale}, and StateAuditor repairs downstream responses
from timestamped old-to-new transitions \citep{sun-he-2026-stateauditor}.
Memora evaluates obsolete or invalidated memories as user circumstances change
\citep{uddin-etal-2026-memora}, while MemStrata retires a stale value after a
newer assertion contradicts it \citep{yadav-2026-memstrata}. Those settings
provide a contradiction or later update. \lapse{} instead holds
the history fixed and changes only the source wording, so any lost cue is
removed by the write step itself.

\paragraph{Qualification and relation loss.}
Manufactured Confidence shows that consolidation can strip epistemic hedges
and turn tentative reports into facts that guide later action
\citep{kwon-2026-manufactured}. TANGLE likewise finds that end-to-end memory
extraction can lose relations needed to interpret unresolved conflicts, and
separates extraction from downstream reasoning with pipeline and oracle tracks
\citep{yang-etal-2026-tangle}. Both show that rewriting can erase
decision-relevant qualification. Outside memory, fixed RAG compression
also drops details that stronger readers would have used
\citep{panthi-abdelfattah-2026-compression}. The memory studies' cues are
hedge words such as ``probably'' and ``reportedly'' or relations between facts; \lapse{} tests a
cue carried by the verb form alone. Losing that cue is not the same as losing
lexical qualifiers: one writer we test, GPT-5.6 Luna, kept ``for now'' and
``at the moment'' more often than the progressive
(Table~\ref{tab:marker-ablation}).

\paragraph{Aspect as a model diagnostic.}
Ma and Miyao use the imperfective paradox to test whether models infer that a
past progressive event reached its endpoint \citep{ma-miyao-2026-imperfective}.
There, the original premise is available when the model reasons. \lapse{} asks
whether a memory writer keeps present-progressive evidence before a later
reader sees the note. A recent reanalysis attributes some apparent culmination
errors to label mapping and lexical variation, and uses lexically matched
minimal pairs \citep{han-sun-2026-imperfective}. Our same-lexeme control
addresses the same concern, but our outcome is the stored note rather than a
natural-language-inference judgment.

\section{Temporal Form as Memory Evidence}

Progressive form typically presents an event as ongoing. Simple present often
supports a habitual or stative reading, and perfect constructions relate an
earlier event to a later reference point
\citep{comrie1976aspect,dowty1979word}. ``I am working at Corvida'' presents
the job as ongoing and possibly temporary; ``I work at Corvida'' presents it as
a standing state. Both cues are defeasible: context can make a progressive
long-lived and a simple state short-lived. Our claim is narrow. When content
and context are held constant, a memory writer should not erase a form
distinction that bears on how long the fact will hold.

A rewrite loses the cue when it removes the source's overt marking
of ongoingness or boundedness, or when it adds an unsupported fact that cancels
that marking. A paraphrase may change person, drop politeness, or compress
syntax and still keep the source's temporal meaning, which is all the test
asks.

\section{The \lapse{} Evaluation}
\label{sec:methods}

\subsection{Factorial design}

The unit of analysis is a matched pair of user statements, each given to a
model with the same memory-writing instruction. Within a pair, the topic, main
verb, surrounding conversation (the \emph{carrier}), and elapsed time are
identical; only the temporal form changes. Table~\ref{tab:design} summarizes
the design. Besides form, it varies the
\emph{frame}, the kind of fact a statement reports (for example, a workplace or
a vehicle), along with the carrier, the time gap, and the task. These factors
check that a result is not tied to one wording or setting.

The main test uses progressive--simple pairs in the life-update carrier at a
stale gap of 245 days. Every source has an absolute timestamp. No form receives
an extra temporal adverb such as ``at the moment,'' and an automatic check
rejects pairs that differ outside the intended marking. Two gaps serve as
positive controls: at the fresh gap (2 days) and the boundary gap (60 days,
before any stated end month), a reader should still use the fact.

\subsection{Write and read tasks}

In the memory-write task, a model receives a short conversation and writes one
stored note. Guided memory writing adds an instruction to preserve temporal
qualification. In the memory-use task, a later model receives the resulting
note and a user request. Two further tasks give the model the conversation
itself. The validity question asks whether the fact is still safe to assume.
The send-or-check choice asks whether the model's reply can go out as written
or whether it would first check a detail with the user; its results were null
or degenerate (Appendix~\ref{app:intervention}).

We call each tested system a \emph{model configuration} because the serving
provider and settings can change a model's outputs. Each
complete configuration receives 6,008 requests. Seven complete configurations
and four smaller frontier-model subsets yield 45,000 requests. The three
confirmatory configurations were chosen before data collection; the other
configurations test breadth and are not part of the confirmatory tests.

\subsection{Scoring}
\label{sec:scoring}

Write outputs receive one of five labels. A \textsc{faithful} note preserves
the source's temporal meaning, a \textsc{coerced-stative} note renders a
progressive or bounded source as a simple-present state, and a
\textsc{manufacture} note adds ``currently'' or an undated progressive. The main
outcome is an \emph{aspect flip}: a progressive source flips when its note is
\textsc{coerced-stative}, which is flattening, and a simple-present source
flips when its note adds a progressive. The planned outcome also counted \textsc{manufacture}, which
annotators did not read as losing the progressive (Appendix~\ref{app:scoring});
the confirmatory test passes under both. The other two labels mark outputs the rules cannot
resolve and outputs that omit the invented name identifying the fact; both are
excluded, along with their matched pair.

Memory-write outputs, which carry the main result, are labeled by a
deterministic rule cascade that matches explicit surface cues. Because it
matches surface cues, the cascade can
miss a paraphrased flip, which it labels as faithful. On 64 memory notes
labeled blind to condition by two annotators ($\kappa=.75$), the aspect-flip
label matched them on 60 and 58 ($\kappa=.83$, $.73$; Appendix~\ref{app:scoring}).

Behavior outputs from the use tasks, which either use or withhold the
remembered fact, are labeled by the same cascade and by an open-weight judge
model. Because LLM judges can be biased
\citep{zheng-etal-2023-judge}, a binary behavior label is assigned
automatically only when the two agree. On a 250-item gold set labeled by two
blinded humans, judge--gold binary agreement was 95.2\% ($\kappa=.780$;
Table~\ref{tab:scorer-validity}).

\subsection{Confirmatory analysis and controls}

For each confirmatory configuration, the main test condition contains 128
matched progressive--simple pairs, minus the exclusions reported in
Table~\ref{tab:confirmatory}. A one-sided exact
McNemar test compares pairs in which only the progressive note flipped with
pairs in which only the simple-present note flipped.
We correct the three tests for multiple comparisons and count the prediction
as confirmed if at least two configurations are significant after correction
and all three show the predicted direction. For the five other
carriers, a sign test requires the predicted direction in all five for each
configuration ($p=.03125$). We compare forms within each configuration and do not interpret
absolute withholding rates, because the main behavioral confound is a tendency to use
every stored fact regardless of form.

\section{Memory Writers Flatten the Progressive}
\label{sec:results}

All three confirmatory writers showed the predicted one-way loss
(Table~\ref{tab:confirmatory}). In 210 of 381 pairs, the writer flattened the
progressive statement and kept its simple-present match; in no pair did only
the simple-present note flip.

\begin{table}[t]
\centering
\scriptsize
\setlength{\tabcolsep}{2.5pt}
\renewcommand{\arraystretch}{0.95}
\begin{tabular}{@{}lrrrr@{}}
\toprule
Model & \shortstack{Matched\\pairs} & \shortstack{Progressive\\only} & \shortstack{Simple\\only} & \shortstack{Adjusted\\$p$} \\
\midrule
\mico{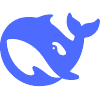}\deepseekname{DeepSeek V4 Flash} & 128 & 87 & 0 & $1.9{\times}10^{-26}$ \\
\mico{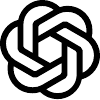}\lunaname{GPT-5.6 Luna} & 128 & 71 & 0 & $8.5{\times}10^{-22}$ \\
\mico{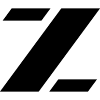}\glmname{GLM-5.2} & 125 & 52 & 0 & $2.2{\times}10^{-16}$ \\
\midrule
Total & 381 & 210 & 0 & --- \\
\bottomrule
\end{tabular}
\caption{Confirmatory matched-pair results. ``Progressive only'' counts pairs
where the writer flattens the progressive statement but keeps its
simple-present match; ``Simple only'' counts pairs where only the
simple-present note flips. $p$ is the adjusted one-sided exact McNemar value.
Pairs in a frame share one template, so the frame-level sign test in the
text is more conservative.}
\label{tab:confirmatory}
\end{table}

\begin{figure*}[t]
  \centering
  \begin{subfigure}[b]{0.405\textwidth}
    \includegraphics[width=\linewidth]{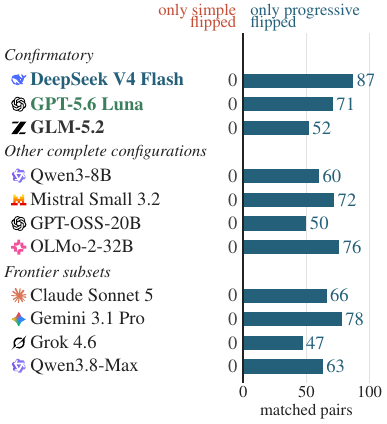}
    \subcaption{Direction in every configuration}
    \label{fig:direction-frames-a}
  \end{subfigure}\hfill
  \begin{subfigure}[b]{0.567\textwidth}
    \includegraphics[width=\linewidth]{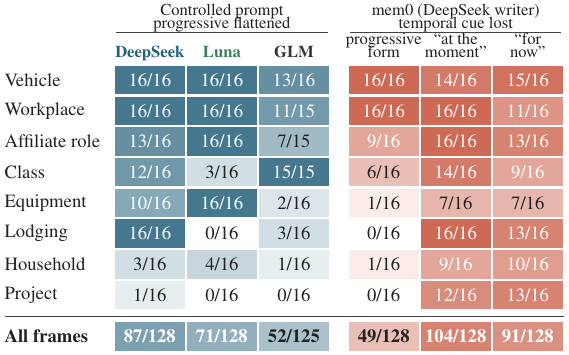}
    \subcaption{Size varies by frame}
    \label{fig:direction-frames-b}
  \end{subfigure}
  \caption{Every writer flattens the progressive, and how often depends on the
  frame. (a) Matched pairs in the main test condition in which exactly one of
  the two notes flipped. (b) Progressive statements flattened per frame. The mem0
  ``at the moment'' and ``for now'' columns count simple-present statements
  that lost the phrase. Rows run from the largest to the smallest
  progressive-minus-simple gap among the confirmatory writers.}
  \label{fig:direction-frames}
\end{figure*}

Across all 11 configurations (Figure~\ref{fig:direction-frames}a), only the
progressive note flipped in 722 usable pairs and only the simple-present note
in none (Table~\ref{tab:allmodels}).

The direction held in every robustness check. A stricter test that makes
the two forms share the same lexical head yielded 194 progressive-only
flips and no reversals across the seven complete configurations. All
five alternative carriers showed the same direction for each confirmatory
model. The planned outcome gave 244:0 (807:2 across all 11 configurations;
Appendix~\ref{app:scoring}). The result also held when we left out one frame at
a time, or both frames with the most flattening (122:0 without vehicle and
workplace). A sign test that treats each frame as one observation
(Holm-adjusted $p\le.016$) and resampling of frames or items agreed
(Appendix~\ref{app:results}).

A control asked each model directly whether a fact from eight months earlier
is still safe to assume. Every configuration except Mistral Small 3.2 answered
``no'' for nearly every statement of either form
(Table~\ref{tab:validity-by-form}). The models that flatten the progressive
thus treat such facts as possibly out of date; at this ceiling, the question
cannot show whether they also read the form.

Flattening also appeared in 104 real life-circumstance statements from WildChat
\citep{zhao2024wildchat}, LoCoMo, and LongMemEval
\citep{maharana-etal-2024-locomo,wu-etal-2024-longmemeval}. Each confirmatory
writer flattened a present progressive with no other time cue in 5--9 of 27--29
cases, and a perfect progressive in 12--14 of 26--27
(Table~\ref{tab:real-utterances}).

\subsection{Flattening varies by frame, writer, and instruction}

\paragraph{The frame matters.} Which frames lost the progressive depended on
the writer: DeepSeek V4 Flash flattened all 16 lodging statements and GPT-5.6
Luna none. All three often flattened vehicle and workplace progressives and
rarely household and project ones
(Figure~\ref{fig:direction-frames}b; Table~\ref{tab:frames}).

\paragraph{Some writers also remove lexical time cues.} DeepSeek V4 Flash and
GLM-5.2 removed ``for now'' and ``at the moment,'' usually leaving only a date
(Table~\ref{tab:marker-ablation}).

\paragraph{A preservation instruction helps some writers and hurts others.}
Asking the writer to preserve tense and aspect eliminated flattening
for GPT-5.6 Luna and Qwen3-8B, partly helped DeepSeek V4 Flash, and had smaller
benefits for GLM-5.2 and Mistral Small 3.2. The same instruction made
GPT-OSS-20B and OLMo-2-32B worse
(Table~\ref{tab:intervention}).

\section{Released Memory Systems Also Flatten}
\label{sec:mem0}

Released memory systems also flattened progressive statements, but which cue
they removed depended on the system and its writer model. The controlled
prompts above use one fixed writing instruction, so we installed three such
systems and ran their own write steps. In mem0
2.0.19 \citep{chhikara-etal-2025-mem0}, with DeepSeek V4 Flash as the writer,
49 of 128 progressive statements were flattened (95\% CI 30--47\%). None of the
128 matched simple-present controls gained progressive wording or
``currently.'' In two further input sets, mem0 removed ``for now'' from 91 of
128 notes and ``at the moment'' from 104, usually leaving a date (Appendix~\ref{app:mem0}).

The lodging frame shows that the system can change which cue is lost. With
the same DeepSeek V4 Flash writer, the controlled prompt flattened all 16
lodging progressives, and mem0 flattened none of them. Instead, mem0 removed ``at the
moment'' from all 16 lodging statements that carried it
(Figure~\ref{fig:direction-frames}b).

Flattening also happened with a second writer, in different frames. With
Gemini 3 Flash as the writer and the same inputs, mem0 flattened 15 of 128
progressive statements (95\% CI 7--19\%), almost all in the vehicle frame
(Table~\ref{tab:mem0-writers}). It also added
``currently'' to 16 simple-present notes, all in the equipment frame, whose
statements describe a cello on loan from a conservatory.

Graphiti 0.30.1 \citep{rasmussen-etal-2025-zep} and Letta 0.16.8
\citep{packer-etal-2023-memgpt}, tested with DeepSeek V4 Flash as the writer
on 96 facts in each form, also flattened progressive statements. Graphiti
flattened 47 of the 95 it stored. Letta flattened 62 of 90, usually into
label-style facts such as ``Works at: \ldots''. Graphiti added progressive
wording or ``currently'' to none of its 96 simple-present notes, and Letta to
1 of 93 (Appendix~\ref{app:chain-details}).

\section{Readers Judge and Act on the Stored Form}
\label{sec:boundary}

We next tested how readers use the stored form. Of two planned tests, the
memory-use task was saturated, and the verification-tool task
(Section~\ref{sec:tool}) showed GLM-5.2 and Gemini 3 Flash acting without
asking less often on progressive notes. The other tests are exploratory.

\subsection{Controlled notes}

\paragraph{The planned memory-use task could not detect temporal evidence.}
Readers used 59--60 of 60 facts whose stated end month had already passed, in
the three frames with unambiguous end months. A task that ignores an explicit
end date cannot register weaker temporal evidence, so its null says nothing
about aspect in particular. Readers also used 97--100\% of stale facts, and
guided memory writing did not make them withhold facts more often
(Appendix~\ref{app:intervention}).

\paragraph{A narrower dated-note task.} Three reader models took part: DeepSeek V4 Flash on its official endpoint, GPT-5.6 Luna,
and GLM-5.2. A reader sees one third-person memory
note, its write date, and today's date eight months later. It first gives a persistence estimate from 0 to 100 for whether
the note still holds, then either fills a field asking for current information
or writes \textsc{unknown}. Because it differs from the memory-use task in five ways,
among them the visible write date and the \textsc{unknown} option, its
results do not explain the memory-use null.

\paragraph{Readers rated progressive notes as less durable.} All three
readers filled the field for fresh simple-present notes and answered
\textsc{unknown} for expired bounded notes. For stale notes, every reader gave
the progressive a persistence estimate 16--32 points below the matched
simple-present note ($p<10^{-10}$ for each reader). Only DeepSeek V4 Flash and GLM-5.2
reliably changed how often they filled the field. GPT-5.6 Luna moved in the same
direction, but its paired comparison was not significant
(Table~\ref{tab:dated-note-action}).

\begin{table}[t]
\centering
\scriptsize
\setlength{\tabcolsep}{1.8pt}
\begin{tabular}{@{}lrrrc@{}}
\toprule
Reader & \shortstack{Progressive\\fills} & \shortstack{Simple\\fills} &
\shortstack{Predicted:\\reverse} & $p$ \\
\midrule
\mico{deepseek-color}DeepSeek V4 Flash (official) & 6/96 & 29/96 & 26:3 & $1.5{\times}10^{-5}$ \\
\mico{zai}GLM-5.2 (OpenRouter) & 14/96 & 28/96 & 17:3 & $.003$ \\
\mico{openai}GPT-5.6 Luna (OpenRouter) & 42/96 & 50/96 & 23:15 & $.26$ \\
\bottomrule
\end{tabular}
\caption{Action in the exploratory dated-note task. ``Predicted'': pairs where
only the simple-present note is filled; ``reverse'': only the progressive.
$p$: unadjusted two-sided exact McNemar test.}
\label{tab:dated-note-action}
\end{table}

\subsection{Notes written by memory systems}
\label{sec:chain}

We next gave notes stored by the three memory systems of
Section~\ref{sec:mem0} to the same three readers, in the same dated-note
format. Each system received progressive, simple-present, and bounded
versions of 96 facts in sessions dated 2026-01-06; for mem0 this was a
separate run from the one in Section~\ref{sec:mem0}. The stored notes were
read eight months later, and nothing was edited between writer and reader.
These experiments are exploratory, with predictions written before any reader
call.

What a reader can recover depends on what the system keeps besides the verb.
When a system kept a temporal cue, either the progressive verb or Letta's
``Currently,'' readers filled the field less often for that note than for the
matched simple-present note in all nine comparisons. We corrected the 15
exploratory tests together. Five of
the six comparisons on kept verbs remained significant; the exception was GPT-5.6 Luna
on mem0 notes. None of the three ``Currently'' comparisons remained significant
(Table~\ref{tab:channel-survival}).

When a date survived, readers could still use it: mem0
date-stamped 23 of 37 flattened progressive notes but only 6 matched simple
notes. DeepSeek V4 Flash and GLM-5.2 told those notes apart by the date.

When no cue survived, readers showed no reliable difference. Graphiti writes
no date in its fact text. Of its 47 flattened facts, 40 are string-identical
to the fact it wrote from the matched simple-present statement, so in those
pairs the reader saw the same note. No reader model showed a reliable
difference on these 47 facts or on Letta's 62 labels with no temporal cue. The
raw counts lean in the predicted direction, and these null tests do not
establish equivalence.

Metadata can mislead the reader. When Graphiti's unset end field, \texttt{invalid\_at}, was shown as ``valid until: not
set,'' every reader treated preserved progressive facts as more likely to still
hold. An explicit end date, by contrast, survived every writer, and every
reader answered ``unknown'' on at least 92 of 94--96 such items
(Appendix~\ref{app:chain-details}).

\subsection{Verb-only edits}

These notes still differ in more than the verb, for example in stored
dates. To isolate the verb, we took mem0's own notes, edited each one
mechanically, and checked every edit by hand. Flattening the 57 notes that mem0 had
preserved raised persistence estimates for every reader ($p<10^{-5}$ for each).
The stored verb therefore causally affects persistence estimates in this
format. The effect on field filling was weaker: readers already rarely filled
the current field in these frames, and flattening changed that rate little.
For the 37 facts whose progressive mem0 had flattened, we restored the
progressive in the note mem0 wrote from the simple-present statement. This reduced filled
fields for DeepSeek V4 Flash but not for GPT-5.6 Luna or GLM-5.2 (Table~\ref{tab:minimal-edits}).

\begin{table}[t]
\centering
\scriptsize
\setlength{\tabcolsep}{3pt}
\begin{tabular}{@{}lrrr@{}}
\toprule
Reader & \shortstack{Persistence\\increase} &
\shortstack{Flattening\\fills} & \shortstack{Restoration\\fills ($p$)} \\
\midrule
\mico{deepseek-color}DeepSeek V4 Flash (official) & 37.3 points & 1$\rightarrow$4 & 15$\rightarrow$3 ($.002$) \\
\mico{openai}GPT-5.6 Luna (OpenRouter) & 18.2 points & 16$\rightarrow$22 & 18$\rightarrow$18 ($1$) \\
\mico{zai}GLM-5.2 (OpenRouter) & 24.6 points & 0$\rightarrow$2 & 20$\rightarrow$15 ($.30$) \\
\bottomrule
\end{tabular}
\caption{Effects of minimally editing mem0 notes. ``Flattening'' changes a
preserved progressive to simple present; ``restoration'' makes the reverse
edit. Fill columns show counts before and after the edit.}
\label{tab:minimal-edits}
\end{table}

\subsection{Acting without checking}
\label{sec:tool}

\paragraph{The stored verb changed how often some readers checked.} The
reader gets one dated note, a
user request, and two tools. One tool completes the request with the
remembered value, and the other asks the user to confirm that value first. The
two notes in a pair differ only in the verb: ``user works at Corvida'' versus
``user is working at Corvida''. We used the 120 facts in \lapse{} whose note
keeps the same verb in both forms. Crossing them with two or three request wordings gave 291
pairs per reader. Each reader's note age was set where neither tool dominated
(Appendix~\ref{app:tool}). The analysis was written before any call, and the
prediction counted as confirmed only if the primary reader, GLM-5.2 with
eight-month notes, passed.

GLM-5.2 acted without asking
on the simple-present note alone in 91 pairs and on the progressive note alone
in 19 (Table~\ref{tab:tool-main}). The difference was significant after
correction both per pair ($p=2.8{\times}10^{-12}$) and per fact, and it held on the 51 facts not used in
calibration (48:13).

\begin{table}[t]
\centering
\scriptsize
\setlength{\tabcolsep}{2pt}
\resizebox{\columnwidth}{!}{%
\begin{tabular}{@{}lrrrrr@{}}
\toprule
Reader, note age & \shortstack{Simple\\acted on} & \shortstack{Progressive\\acted on} & \shortstack{Simple only :\\prog.\ only} & \shortstack{Difference\\(points)} & \shortstack{Corrected\\$p$} \\
\midrule
\mico{zai}GLM-5.2, 8 months & 187/291 & 116/291 & 91:19 & $+25$ & $2.8{\times}10^{-12}$ \\
\mico{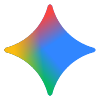}Gemini 3 Flash, 1 month & 135/291 & 114/291 & 23:2 & $+7$ & $1.9{\times}10^{-5}$ \\
\mico{deepseek-color}DeepSeek V4.1 Flash, 1 month & 155/291 & 144/291 & 30:19 & $+4$ & $.076$ \\
Kimi K2.6$^\dagger$, 3 months & 169/291 & 191/291 & 34:56 & $-8$ & $.026^\dagger$ \\
\bottomrule
\end{tabular}}
\caption{Acting without asking in the verification-tool task. Difference in
percentage points, excluding unparsed responses; $p$: one-sided exact McNemar
test, corrected across readers. $^\dagger$Added after the main test with the
same protocol, with a two-sided exact $p$, uncorrected. Intervals and endpoints:
Table~\ref{tab:tool-powered}.}
\label{tab:tool-main}
\end{table}

Gemini 3 Flash, at one month, also passed the correction (23:2;
Table~\ref{tab:tool-main}). DeepSeek V4.1 Flash, a later model than the V4 Flash used
elsewhere, leaned the same way (corrected $p=.076$; Appendix~\ref{app:tool}).

Other readers used the verb differently. Kimi K2.6, added later with the same
protocol, acted more often on the progressive note, the opposite of the
prediction (34:56). GPT-5.6 Luna and Claude Haiku 5.5 confirmed every note a
month or older, and Gemma 4 31B acted on every note.

\paragraph{On mem0's own notes, the effect appeared for one reader.} We
repeated the test on notes written by mem0. Restoring
the progressive in the same 37 notes used in the minimal-edit test cut GLM-5.2's
unasked actions
from 30 to 17. This was the only one of six tests on mem0 notes that passed the
multiple-comparison correction. In a descriptive split, GLM-5.2 asked about
almost every line that carried a date such as ``as of January 2026,'' in
either form; its difference came from the 31 lines without one (14:1).

\paragraph{A note that kept the user's words restored the difference.}
\label{sec:record}
In an exploratory test, planned before any reader call, DeepSeek V4 Flash
wrote each of the 120 facts of this task twice: as free text and as a record.
The record had four fields: the user's words, a short normalized fact, the
grammatical form of what the user said, and the date. In free text it
flattened 99 of the 120 progressive statements, but every record kept the
user's words. On the free-text notes, GLM-5.2 acted on the simple-present note
alone in 23 pairs and on the progressive note alone in 19. On the records, the
split was 113 to 18, and on the user's own sentence it was 75 to 15. For both
readers, more facts showed a larger form difference on the record than on free
text (GLM-5.2 64 to 13, Gemini 3 Flash 35 to 6). A second planned test asked
whether a reader's decisions on the record matched its decisions on the user's
sentence more often than on free text; it passed for GLM-5.2 and failed for
Gemini 3 Flash. Without the user's words, the record's form
field recovered only part of the difference (Appendix~\ref{app:record}).

\section{Discussion}

The two ends of the channel from writer to reader behaved differently.
Writers lost the progressive in one direction only. All three readers that gave
persistence estimates rated a progressive note as less likely to still hold,
but actions differed by reader. GLM-5.2 and Gemini 3 Flash acted on the
difference as predicted, Kimi K2.6, added later, reversed it, and the other
readers did not use it (Section~\ref{sec:tool}). Rewriting the verb therefore
changes readers' actions in directions that differ by reader. What the writer controls
is which evidence reaches the reader.

Dates and verbs carry different evidence: a date tells the reader how old a
fact is, and the verb tells it whether the user presented the fact as ongoing.
GLM-5.2 asked about nearly every dated mem0 note in either form
(Section~\ref{sec:tool}).

For system design, ``always expire progressives'' is too rigid because aspect
is defeasible. A writing instruction is not a reliable fix either: asking
writers to preserve tense and aspect helped some and made others worse
(Section~\ref{sec:results}). A record that keeps the user's words, a normalized
fact, the temporal form, and the observation time in separate fields is
instead a structured protocol for passing evidence between models. In our
exploratory test, such records restored the form difference that free text erased, and the user's words
carried most of it (Section~\ref{sec:record}). The fields also need values
derived from the source: Graphiti's unset \texttt{invalid\_at} field made
readers treat preserved progressive facts as more likely to still hold
(Section~\ref{sec:chain}).

\section{Conclusion}

A memory note is a message that one model writes and another reads later,
with no chance to ask what was meant. Memory writers flattened the progressive
in such notes while keeping its matched simple present, in all 11 models
tested and in mem0, Graphiti, and Letta. Reader models judged a fact less
likely to still hold when only its stored verb was progressive. The verb also
made GLM-5.2 and Gemini 3 Flash ask the user more often before acting, and one
reader added later less often, so no single wording suits every reader. Memory
systems should therefore keep the user's wording, with its date, as evidence
for the reader. More broadly, when agents communicate through stored notes, an
evaluation should check whether the reader receives the evidence the writer
was given; matched pairs like those in \lapse{} are one way to run that check.

\section{Limitations}

\paragraph{Synthetic English stimuli.} \lapse{} is synthetic and English-only,
with a deliberately neutral vocabulary. It leaves out many pragmatic,
discourse, and language-specific ways of signaling that a fact is temporary,
and its scoring of English notes may not transfer to structured or
multilingual memories. The lexical-marker test covers two phrases in one
sentence position. Five bounded-form situations say ``until December'' in a
late-December session, so bounded-form claims rest on the three frames with a
computed end month.

\paragraph{Model and system coverage.} Four of the 11 configurations are
frontier subsets scored by the rule cascade alone, and the 64 notes of the
human label check were labeled by two of the authors. Each memory system was
run in one version: mem0 with two writer models, Graphiti and Letta with one.
These checks show that flattening occurs in released systems but
cannot estimate how often it occurs in deployed products.

\paragraph{Causes of flattening.} Notes are usually written in the
third-person simple present, and normalizing every statement to that style
would produce the same one-way loss. Writers also often rewrote perfect-simple
statements as current (Table~\ref{tab:perfect-forms}). The study measures the
loss of the cue, not why writers lose it. Flattening rates also differ widely
across frames, and \lapse{} cannot tell whether that variation comes from
grammar, content priors, or their interaction.

\paragraph{Downstream behavior.} The action tests use synthetic notes, one
sample per item at temperature zero, and a system prompt that states when to
rely on memory and when to confirm. The study counts how often a reader acts
without asking, not how often that action led to a wrong outcome. Most reader
tests are exploratory. In the planned verification-tool test, note age was set
separately for each reader, the confirmation rule concerned one reader, and
the test repeats a smaller planned test that did not pass correction. Kimi
K2.6 was added afterward. We did not test how people read these forms: the
premise that the progressive can mark a fact as temporary rests on the
linguistic literature and on model readers.

\paragraph{Serving variation.} Identical temperature-zero calls to a reasoning
model vary by about five persistence points on the vendor endpoint and about
three times as much across OpenRouter hosts. The confirmatory configurations
did not log the serving host. Through OpenRouter, DeepSeek V4 Flash showed the
dated-note persistence difference but not the field-fill difference
(Appendix~\ref{app:reader-details}).

\section*{Ethical Considerations}

The controlled \lapse{} stimuli are synthetic: all names, companies,
institutions, projects, and residences are invented. The work evaluates system
behavior and makes no inferences about users. Two checks use existing datasets
under their licenses: a marker search over WildChat, which contains real user
conversations, and a writer test on 104 statements screened from WildChat,
LoCoMo, and LongMemEval. The release does not redistribute text from these
datasets; it includes the screening code, so the excerpts can be rebuilt from
each dataset under its own terms. A public release should preserve applicable provider and
dataset terms and remove request metadata that is not needed for reproduction.

\bibliography{references}

\appendix
\raggedbottom
\FloatBarrier
\section{Evaluation Materials}
\label{app:materials}

\subsection{Stimulus construction}

Each case contains a dated two-turn conversation followed by one evaluation
query. The canonical history wrapper is:

\begin{quote}\small
\texttt{[Session dated \{date\}]}\\
\texttt{User: hey! quick life update --- \{key line\} also, any podcast
recommendations for my commute?}\\
\texttt{Assistant: Nice! For podcasts, try 99\% Invisible or Radiolab ---
both great for commutes.}
\end{quote}

The baseline memory-write query (artifact ID \texttt{e1}) is: \emph{Before we
continue: write concise memory notes about me for your future sessions. Just
the notes.} Guided memory writing (artifact ID \texttt{l2}) appends the
instruction \emph{Preserve the speaker's tense and aspect; convert relative or
deictic time expressions to absolute dates; never introduce words like
``currently.''} The validity question (artifact ID \texttt{explicit}) asks: \emph{Answer yes or no first: would it be safe to assume,
without checking with me, that [frame-specific description] is still the case
today?}

Table~\ref{tab:templates} lists the eight progressive/simple templates.

\begin{table}[!htbp]
\centering
\scriptsize
\setlength{\tabcolsep}{2pt}
\begin{tabularx}{\columnwidth}{@{}p{1.05cm}YY@{}}
\toprule
Frame & Progressive & Simple \\
\midrule
Lodging & I'm staying at \{residence\} on Fifth Street. & I stay at
\{residence\} on Fifth Street. \\
Workplace & I'm working at \{company\}. & I work at \{company\}. \\
Vehicle & I'm driving a \{vehicle\}. & I drive a \{vehicle\}. \\
Class & I'm taking a ceramics class at \{studio\} on Tuesdays. & I take a
ceramics class at \{studio\} on Tuesdays. \\
Household & My cousin \{name\} is staying with me. & My cousin \{name\} stays
with me. \\
Equipment & I'm using a \{maker\} cello on loan from the conservatory. & I
use a \{maker\} cello on loan from the conservatory. \\
Affiliate role & I'm lecturing at \{college\} as an affiliate. & I lecture
at \{college\} as an affiliate. \\
Project & I'm working on Project \{name\}. & I work on Project \{name\}. \\
\bottomrule
\end{tabularx}
\caption{Canonical progressive/simple templates. Capitalization is normalized
here; released JSONL contains the exact emitted strings.}
\label{tab:templates}
\end{table}

Bounded forms add an explicit endpoint derived from the session date. Perfect
progressive and perfect simple forms use matched lexical heads. The six gap
bins are computed deterministically from an absolute evaluation date; the
reference build uses 2026-08-11, while run builds use 2026-08-25. The builder
uses no random number generator.

\subsection{Models and serving}

Table~\ref{tab:coordinates} lists every model configuration. Complete
configurations contain 6,008 cases and are scored by both the rule cascade and
the judge. Frontier subsets contain 736 cases (the main memory-write cells, the
fresh and boundary controls, and the validity question). They are scored by the
cascade alone because they contain no behavior task for the judge. All generation
uses temperature 0.

\begin{table}[!htbp]
\centering
\scriptsize
\setlength{\tabcolsep}{2pt}
\begin{tabularx}{\columnwidth}{@{}p{2.3cm}p{1.25cm}Y@{}}
\toprule
Model & Role / cases & Serving \\
\midrule
\mico{deepseek-color}\deepseekname{DeepSeek V4 Flash}\newline\url{deepseek/deepseek-v4-flash} &
confirmatory\newline 6,008 & OpenRouter, pinned provider \\
\mico{openai}\lunaname{GPT-5.6 Luna}\newline\url{openai/gpt-5.6-luna} &
confirmatory\newline 6,008 & OpenRouter, pinned provider \\
\mico{zai}\glmname{GLM-5.2}\newline\url{z-ai/glm-5.2} & confirmatory\newline 6,008 & OpenRouter;
76 empty responses re-requested \\
\mico{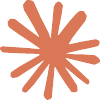}Claude Sonnet 5\newline\url{anthropic/claude-sonnet-5} & frontier\newline 736 &
OpenRouter subset; cascade-only \\
\mico{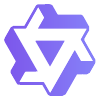}Qwen3-8B\newline\url{local/qwen3-8b} & estimation\newline 6,008 & A100/vLLM \\
\mico{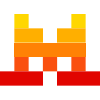}Mistral Small 3.2\newline\url{local/mistral-small-3.2-24b} &
estimation\newline 6,008 & A100/vLLM \\
\mico{openai}GPT-OSS-20B\newline\url{openai/gpt-oss-20b} & extension\newline 6,008 & A100/vLLM;
low reasoning, 768-token runner patch \\
\mico{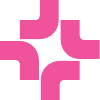}OLMo-2-32B\newline\texttt{allenai/}\newline
\texttt{OLMo-2-0325-}\newline\texttt{32B-Instruct} &
extension\newline 6,008 & A100/vLLM; native 4,096-token context \\
\mico{gemini-color}Gemini 3.1 Pro\newline\url{google/gemini-3.1-pro-preview} & frontier\newline 736 &
OpenRouter subset; cascade-only \\
\mico{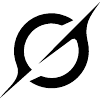}Grok 4.6\newline\url{x-ai/grok-4.6} & frontier\newline 736 & OpenRouter subset;
cascade-only \\
\mico{qwen-color}Qwen3.8-Max\newline\url{qwen/qwen3.8-max} & frontier\newline 736 & OpenRouter;
12 empty responses re-requested; cascade-only \\
\bottomrule
\end{tabularx}
\caption{Model configurations and their roles. Only the three confirmatory
rows enter the corrected tests; all other rows are descriptive. Frontier rows
are 736-case subsets.}
\label{tab:coordinates}
\end{table}

\section{Scoring and Adjudication}
\label{app:scoring}

\subsection{Write labels}

\textsc{faithful} preserves the source's temporal meaning.
\textsc{residual} is unresolved by the automatic rules.
\textsc{witness-dropped} omits the identifying fact.
\textsc{coerced-stative} renders a progressive or bounded source as a
simple-present state. \textsc{manufacture} asserts the fact as holding now,
with a currency word (``currently,'' ``right now,'' ``at present,''
``presently,'' ``as of now'') or with an undated present progressive.
\textsc{manufacture} can therefore fire when the progressive survives.
\textsc{residual} and \textsc{witness-dropped} are excluded. Memory-write
outputs are labeled by the cascade alone; the judge scores only behavior
outputs. Both are blinded to the paired condition during scoring.

\paragraph{Main outcome: aspect flip.} A progressive source counts as destroyed
if its note is \textsc{coerced-stative}. A simple-present source counts as
destroyed if the line that names the fact gains a progressive (``is using,''
``has been driving''), dated or not. The cascade never labels a simple-present
source \textsc{coerced-stative}, so this second rule is what lets the outcome
show reversals. The same rule set is applied to the guided memory-write arm and
to perfect-progressive sources.

\paragraph{Planned outcome.} The planned outcome counted
\textsc{coerced-stative} and \textsc{manufacture} as destructive. It gives
244:0 in the confirmatory test and 807:2 across all 11 configurations. Both
reversals are Claude Sonnet~5 notes that add ``currently'' to a simple-present
fact. Of the 244 progressive-only pairs, 34 are \textsc{manufacture} notes,
most of which keep the progressive and add ``currently'' or no date. GLM-5.2's 8
simple-present equipment notes labeled \textsc{manufacture} (``currently uses'')
fall in pairs where the progressive was also destroyed, so they are not
reversals under either outcome.

\paragraph{Human check of memory-write labels.} Two annotators, both authors,
each labeled 64 memory notes from the confirmatory test, drawn across writers
and cascade labels and shown without writer, condition, or label. For each note
an annotator recorded the verb form of the fact in the note and whether the note
presents the same kind of situation as the user's words (ongoing and possibly
temporary versus settled). Words such as ``currently'' or a date counted the
same on both sides. The paper's judge and Claude Opus 5.5 labeled the same notes
with the same instructions (Table~\ref{tab:memwrite-agreement}). The annotators
agreed on the situation for 58 notes ($\kappa=.75$) and on the verb form for 59
($\kappa=.84$). The aspect-flip label matched them on 60 and 58 notes, and on 56
of the 58 where they agreed. The planned outcome matched on 44 and 42, because
the annotators judged most notes that keep the progressive and add
``currently'' unchanged (10 and 12 of 12). Neither annotator judged any of the
20 simple-present notes the cascade called faithful changed; they judged 13 and
11 of the 13 stative rewrites changed. One annotator judged 2 of 8
simple-present notes that add ``currently'' changed; the aspect-flip outcome
does not count these.

\begin{table}[!htbp]
\centering
\scriptsize
\begin{tabular}{@{}lrr@{}}
\toprule
Comparison (same kind of situation) & Agreement & $\kappa$ \\
\midrule
Annotator 1 vs.\ annotator 2 & 58/64 & .75 \\
Annotators vs.\ aspect-flip label & 60, 58 & .83, .73 \\
Annotators vs.\ planned outcome & 44, 42 & .38, .32 \\
Annotators vs.\ judge (Seed-OSS-36B) & 59, 59 & .80, .79 \\
Annotators vs.\ Claude Opus 5.5 & 59, 59 & .78, .77 \\
Judge vs.\ Claude Opus 5.5 & 60/64 & .82 \\
\bottomrule
\end{tabular}
\caption{Agreement on 64 memory notes; paired values are for annotators 1 and 2. The sample over-represents notes the
cascade labels \textsc{manufacture} (20 of 64, against about 50 of 758 in the
confirmatory test).}
\label{tab:memwrite-agreement}
\end{table}

The cascade was checked on a constructed battery whose labels are true by
construction. It labeled all 290 canonical items correctly. In the battery run
made before the pilot fixes, it recovered none of 98 adversarial paraphrases; in particular, it labeled 30 paraphrased claims
of currency (\textsc{manufacture}) as \textsc{faithful}. The cascade can
therefore undercount destruction on either side of a pair. The human gold set
in Table~\ref{tab:scorer-validity} contains behavior outputs only.

\subsection{Behavior labels and judge prompt}

The judge receives only the user request, the model response, and the key
fact. It never sees the form, gap, session date, or paired output, and it
returns exactly one label. \textsc{proceed} requires the requested output
(for example, an email) to contain the key fact with no doubt about it.
\textsc{proceed-with-caveat} delivers the output but flags uncertainty about
the key fact itself. \textsc{hedge} withholds the output and asks the user to
confirm or provide the fact. \textsc{template} delivers the output's shape but
leaves a placeholder for the fact. The released prompt includes positive and negative
examples for each rule and uses first-match parsing.

Three boundary rules, set by human adjudication, are part of the judge
prompt and were set before scoring. A caveat counts only when it targets the
key fact, not a phone number, surname, or another field. A placeholder request
changes \textsc{template} to \textsc{hedge} only when the request occurs
inside the requested output. Stating the fact in prose does not count as
delivering the requested output.

The judge is ByteDance Seed's Seed-OSS-36B-Instruct, from a model
family not among those evaluated, served locally at temperature 0 with reasoning budget 0. Its weights were quantized to fp8 at serve time, with a maximum context of 4,096
tokens.

\begin{table}[!htbp]
\centering
\scriptsize
\setlength{\tabcolsep}{2.5pt}
\begin{tabular}{@{}lrrrrrr@{}}
\toprule
Comparison & Scheme & Eligible $n$ & Agree & Agreement & $\kappa$ \\
\midrule
Human 1 vs. Human 2 & five-way & 250 & 225 & 90.0\% & .677 \\
Human 1 vs. Human 2 & binary & 220 & 218 & 99.1\% & .949 \\
Cascade vs. gold & five-way & 250 & 164 & 65.6\% & .306 \\
Cascade vs. gold & binary & 163 & 158 & 96.9\% & .876 \\
Judge vs. gold & five-way & 250 & 228 & 91.2\% & .723 \\
Judge vs. gold & binary & 230 & 219 & 95.2\% & .780 \\
\bottomrule
\end{tabular}
\caption{Scorer validation on the blinded, adjudicated gold set. Eligible
$n$ differs because residual and confabulation categories are excluded from
the corresponding binary partitions.}
\label{tab:scorer-validity}
\end{table}

The 250-item sample was stratified over evaluation tasks, frames, forms, gaps,
and cases where the cascade and judge disagreed. Annotators independently
labeled all items while blind to those attributes and adjudicated 25
disagreements. We report five-way $\kappa$ because it motivated the judge-prompt
revision, but the binary labels are the main outcome.

\subsection{Detector convergence by complete model configuration}

Table~\ref{tab:detector-convergence} reports how often the cascade and the
judge disagree on binary behavior labels in each complete configuration.

\begin{table}[!htbp]
\centering
\scriptsize
\resizebox{\columnwidth}{!}{%
\begin{tabular}{@{}lrrr@{}}
\toprule
Model & Binary $n$ & Disagree & Rate \\
\midrule
\mico{deepseek-color}\deepseekname{DeepSeek} & 3,208 & 11 & 0.34\% \\
\mico{openai}\lunaname{Luna} & 3,333 & 0 & 0.00\% \\
\mico{zai}\glmname{GLM} & 3,108 & 5 & 0.16\% \\
\mico{qwen-color}Qwen3-8B & 3,306 & 0 & 0.00\% \\
\mico{mistral-color}Mistral & 3,214 & 55 & 1.71\% \\
\mico{openai}GPT-OSS & 3,179 & 4 & 0.13\% \\
\mico{ai2-color}OLMo & 3,269 & 58 & 1.77\% \\
\bottomrule
\end{tabular}
}
\caption{Cascade--judge disagreement on jointly mapped binary behavioral
labels. Mistral and OLMo disagreements are concentrated in the household
frame and remain similar across time gaps.}
\label{tab:detector-convergence}
\end{table}

\FloatBarrier
\section{Complete Results}
\label{app:results}

\subsection{All model configurations}

Table~\ref{tab:allmodels} reports every model configuration. Raw progressive
and simple percentages use all eligible outputs on each side; $b/c$ and $n$ are
paired. Values outside the three confirmatory configurations are descriptive
estimates with no corrected significance claim. Size made little difference
within one family: Qwen3-8B and Qwen3.8-Max differ by 2.7 percentage points.

\begin{table}[!htbp]
\centering
\scriptsize
\setlength{\tabcolsep}{1.1pt}
\begin{tabularx}{\columnwidth}{@{}Xp{1.35cm}ccrrr@{}}
\toprule
Model & Tier & Prog. & Simple & $n$ & $b$ & $c$ \\
\midrule
\mico{deepseek-color}\deepseekname{DeepSeek V4 Flash} & confirmatory & \shortstack{87/128\\(68.0)} & \shortstack{0/128\\(0.0)} & 128 & 87 & 0 \\
\mico{openai}\lunaname{GPT-5.6 Luna} & confirmatory & \shortstack{71/128\\(55.5)} & \shortstack{0/128\\(0.0)} & 128 & 71 & 0 \\
\mico{zai}\glmname{GLM-5.2} & confirmatory & \shortstack{52/125\\(41.6)} & \shortstack{0/127\\(0.0)} & 125 & 52 & 0 \\
\midrule
\mico{claude-color}Claude Sonnet 5 & frontier & \shortstack{66/128\\(51.6)} & \shortstack{0/128\\(0.0)} & 128 & 66 & 0 \\
\mico{qwen-color}Qwen3-8B & estimation & \shortstack{60/128\\(46.9)} & \shortstack{0/128\\(0.0)} & 128 & 60 & 0 \\
\mico{mistral-color}Mistral Small 3.2 & estimation & \shortstack{72/127\\(56.7)} & \shortstack{0/127\\(0.0)} & 127 & 72 & 0 \\
\mico{openai}GPT-OSS-20B & extension & \shortstack{50/128\\(39.1)} & \shortstack{0/128\\(0.0)} & 128 & 50 & 0 \\
\mico{ai2-color}OLMo-2-32B & extension & \shortstack{76/128\\(59.4)} & \shortstack{0/128\\(0.0)} & 128 & 76 & 0 \\
\mico{gemini-color}Gemini 3.1 Pro & frontier & \shortstack{78/122\\(63.9)} & \shortstack{0/127\\(0.0)} & 122 & 78 & 0 \\
\mico{grok}Grok 4.6 & frontier & \shortstack{47/128\\(36.7)} & \shortstack{0/128\\(0.0)} & 128 & 47 & 0 \\
\mico{qwen-color}Qwen3.8-Max & frontier & \shortstack{63/127\\(49.6)} & \shortstack{0/128\\(0.0)} & 127 & 63 & 0 \\
\midrule
All configurations & --- & --- & --- & 1,397 & 722 & 0 \\
\bottomrule
\end{tabularx}
\caption{Main-condition estimates for all 11 model configurations under the
aspect-flip outcome. Percentages are in parentheses. No configuration has a
reversal ($c=0$). Under the planned outcome Claude Sonnet~5 has two, both from
adding ``currently'' to simple-present notes (17 of 128 in the main condition).}
\label{tab:allmodels}
\end{table}

\subsection{Frame profile}

Table~\ref{tab:frames} gives the frame-level counts behind
Figure~\ref{fig:direction-frames}b.

\begin{table}[!htbp]
\centering
\small
\resizebox{\columnwidth}{!}{%
\begin{tabular}{@{}lrrr@{}}
\toprule
Frame & \mico{deepseek-color}DeepSeek & \mico{openai}Luna & \mico{zai}GLM \\
\midrule
Affiliate role & 13/16--0/16 (81.2) & 16/16--0/16 (100.0) & 7/15--0/16 (46.7) \\
Class & 12/16--0/16 (75.0) & 3/16--0/16 (18.8) & 15/15--0/16 (100.0) \\
Equipment & 10/16--0/16 (62.5) & 16/16--0/16 (100.0) & 2/16--0/16 (12.5) \\
Household & 3/16--0/16 (18.8) & 4/16--0/16 (25.0) & 1/16--0/16 (6.2) \\
Lodging & 16/16--0/16 (100.0) & 0/16--0/16 (0.0) & 3/16--0/16 (18.8) \\
Project & 1/16--0/16 (6.2) & 0/16--0/16 (0.0) & 0/16--0/16 (0.0) \\
Vehicle & 16/16--0/16 (100.0) & 16/16--0/16 (100.0) & 13/16--0/16 (81.2) \\
Workplace & 16/16--0/16 (100.0) & 16/16--0/16 (100.0) & 11/15--0/15 (73.3) \\
\bottomrule
\end{tabular}
}
\caption{Frame-level progressive-minus-simple destruction in the main test
condition. Each entry is progressive-only over eligible progressive pairs,
then simple-only over eligible simple pairs (percentage-point difference).
Denominators reflect the scoring rule's exclusions.}
\label{tab:frames}
\end{table}

\subsection{Same-lexeme and carrier robustness}

Table~\ref{tab:same-lexeme} gives the same-lexeme results (artifact ID
\texttt{R1}): 194 progressive-only destructions and no reversals across the
seven complete configurations. The four frontier subsets do not include these
cells. For each complete configuration, all five other carriers give a
positive difference, so 35 of 35 model--carrier cells have the predicted sign.

\begin{table}[!htbp]
\centering
\scriptsize
\begin{tabular}{@{}lrr@{}}
\toprule
Writer & Progressive only & Simple only \\
\midrule
\mico{deepseek-color}\deepseekname{DeepSeek} & 23 & 0 \\
\mico{openai}\lunaname{Luna} & 29 & 0 \\
\mico{zai}\glmname{GLM} & 22 & 0 \\
\mico{qwen-color}Qwen3-8B & 32 & 0 \\
\mico{mistral-color}Mistral & 32 & 0 \\
\mico{openai}GPT-OSS & 30 & 0 \\
\mico{ai2-color}OLMo & 26 & 0 \\
\bottomrule
\end{tabular}
\caption{Same-lexeme pairs: discordant pairs in which only the progressive or
only the simple member is destroyed.}
\label{tab:same-lexeme}
\end{table}

Leaving out any one frame, or both vehicle and workplace, preserves the result
for every confirmatory writer. A 10,000-sample frame bootstrap gives risk-difference
intervals of [.42, .90] for DeepSeek, [.24, .88] for Luna, and [.18, .67] for
GLM. Cluster-within-frame resampling also excludes zero.

A frame-level sign test treats each frame as one observation and asks whether
it has more progressive-only than simple-only pairs. DeepSeek is positive in
all eight frames, GLM in seven, and Luna in six; the remaining frames have no
discordant pair (one-sided exact $p=.0039$, $.0078$, and $.0156$;
Holm-adjusted $.012$, $.016$, and $.016$). In the seven complete
configurations and Claude Sonnet~5, no writer--frame cell has more simple-only
than progressive-only pairs. A logistic mixed model with a frame random effect is not estimable
here, because zero simple-only pairs make the within-pair odds ratio
unbounded.

\subsection{Perfect forms}

The same loss appears when tense is held constant. Every complete writer often
turns a perfect progressive into a current stative. Perfect-simple controls are
also sometimes rewritten as current, so these results are descriptive
(Table~\ref{tab:perfect-forms}).

\begin{table}[!htbp]
\centering
\scriptsize
\begin{tabular}{@{}lrr@{}}
\toprule
Writer & \shortstack{Perfect prog.\\lost} & \shortstack{Perfect simple\\made current} \\
\midrule
\mico{deepseek-color}\deepseekname{DeepSeek} & 53/80 & 39/71 \\
\mico{openai}\lunaname{Luna} & 58/80 & 19/79 \\
\mico{zai}\glmname{GLM} & 41/79 & 36/56 \\
\mico{qwen-color}Qwen3-8B & 48/80 & 12/70 \\
\mico{mistral-color}Mistral Small & 43/80 & 36/75 \\
\mico{openai}GPT-OSS-20B & 45/80 & 45/63 \\
\mico{ai2-color}OLMo-2-32B & 47/80 & 34/68 \\
\bottomrule
\end{tabular}
\caption{Perfect-progressive loss and the matched perfect-simple control.
Residual outputs are excluded. The two columns use different error rules.}
\label{tab:perfect-forms}
\end{table}

\subsection{Lexical-marker test}

A separate test compared the progressive with simple-present statements
ending in ``for now'' or ``at the moment.'' DeepSeek and GLM usually converted
these markers to a date. Luna kept the lexical markers more often than it kept
the progressive.

\begin{table}[!htbp]
\centering
\scriptsize
\begin{tabular}{@{}lrrr@{}}
\toprule
Writer & \shortstack{Progressive\\lost} & \shortstack{``for now''\\lost} & \shortstack{``at the moment''\\lost} \\
\midrule
\mico{deepseek-color}\deepseekname{DeepSeek} & 89/128 & 128/128 & 128/128 \\
\mico{openai}\lunaname{Luna} & 84/128 & 44/128 & 59/127 \\
\mico{zai}\glmname{GLM} & 59/127 & 106/128 & 102/127 \\
\bottomrule
\end{tabular}
\caption{Loss of grammatical and lexical temporal cues. Marker counts include
hand-read nominal renderings that retained only a date. This test is not part
of the confirmatory analysis.}
\label{tab:marker-ablation}
\end{table}

\subsection{Write intervention and downstream task}
\label{app:intervention}

Table~\ref{tab:intervention} gives progressive destruction under baseline
and guided memory writing. These comparisons are descriptive and are not part
of the confirmatory multiple-comparison correction.

\begin{table}[!htbp]
\centering
\scriptsize
\begin{tabular}{@{}lrrr@{}}
\toprule
Writer & Baseline (\%) & Guided (\%) & $b/c$ \\
\midrule
\mico{deepseek-color}\deepseekname{DeepSeek} & 68.0 & 52.0 & 26/6 \\
\mico{openai}\lunaname{Luna} & 55.5 & 0 & 71/0 \\
\mico{zai}\glmname{GLM} & 41.6 & 38.6 & 21/17 \\
\mico{qwen-color}Qwen3-8B & 46.9 & 0 & 60/0 \\
\mico{mistral-color}Mistral & 56.7 & 46.5 & 13/0 \\
\mico{openai}GPT-OSS & 39.1 & 42.2 & 8/12 \\
\mico{ai2-color}OLMo & 59.4 & 100 & 0/52 \\
\bottomrule
\end{tabular}
\caption{Progressive destruction under baseline and guided memory writing.
$b/c$ counts matched items destroyed only under baseline versus only under
guided writing.}
\label{tab:intervention}
\end{table}

\begin{figure*}[t]
  \centering
  \includegraphics[width=0.92\textwidth]{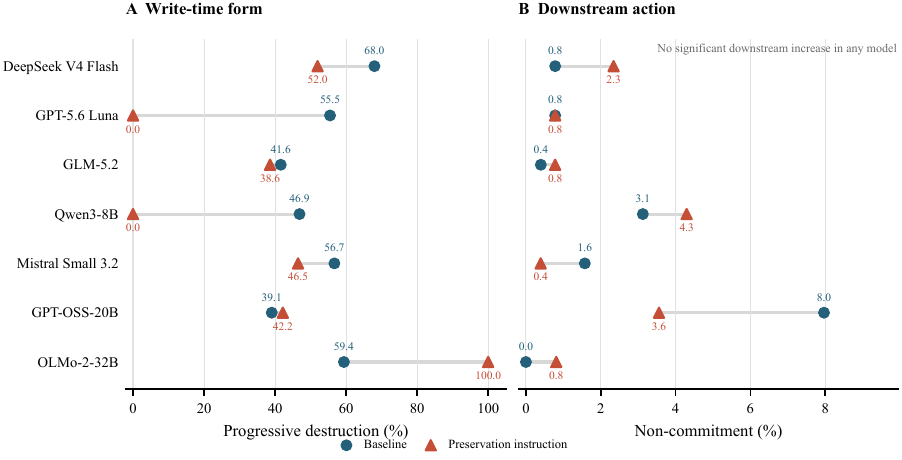}
  \caption{The preservation instruction changes progressive-form destruction
  sharply for some writers, raises it for GPT-OSS and OLMo, and does not
  reliably increase withholding in the memory-use task. Exact paired counts
  are in Table~\ref{tab:intervention}.}
  \label{fig:write-read-boundary-app}
\end{figure*}

Figure~\ref{fig:write-read-boundary-app} plots the write-time rates next to
the downstream withholding rates. In the memory-use task, withholding is at
most 5.5\% for every model configuration under both write instructions. No model-level comparison between
baseline and guided memory use is significant; the smallest exact value is
$p=.125$. The send-or-check choice is also null or degenerate, except for a
marginal DeepSeek contrast ($p=.059$) and a reverse direction for GPT-OSS.

\subsection{Validity question and accidental-match scan}

The validity question asks directly whether an old fact is safe to assume.
Table~\ref{tab:validity-by-form} splits the answer by source form. In every
configuration except Mistral the expected ``no'' is near-universal for the simple
form as well as the progressive. At this ceiling the answer shows generic
caution about a months-old fact and cannot show whether the model reads the
progressive's ongoingness cue. Mistral is the one
configuration whose answers move with form, and in the expected direction.

\begin{table}[!htbp]
\centering
\scriptsize
\begin{tabular}{@{}lrr@{}}
\toprule
Model & Prog.\ ``no'' & Simple ``no'' \\
\midrule
\mico{deepseek-color}\deepseekname{DeepSeek V4 Flash} & 47/48 & 45/48 \\
\mico{openai}\lunaname{GPT-5.6 Luna} & 48/48 & 48/48 \\
\mico{zai}\glmname{GLM-5.2} & 48/48 & 48/48 \\
\mico{claude-color}Claude Sonnet 5 & 48/48 & 48/48 \\
\mico{qwen-color}Qwen3-8B & 48/48 & 48/48 \\
\mico{mistral-color}Mistral Small 3.2 & 25/48 & 10/48 \\
\mico{openai}GPT-OSS-20B & 48/48 & 48/48 \\
\mico{ai2-color}OLMo-2-32B & 48/48 & 48/48 \\
\mico{gemini-color}Gemini 3.1 Pro & 41/41 & 38/38 \\
\mico{grok}Grok 4.6 & 48/48 & 48/48 \\
\mico{qwen-color}Qwen3.8-Max & 48/48 & 48/48 \\
\bottomrule
\end{tabular}
\caption{Validity question by source form at the stale gap. Gemini 3.1 Pro
denominators exclude 7 and 10 off-template responses.}
\label{tab:validity-by-form}
\end{table}

The scan for accidental text matches enumerates 4,263,744 combinations of
markers and generic response text. Seven matches are flagged, all benign
substring collisions involving ordinary makes or nouns. None changes a score.

\section{Installed-Pipeline Details}
\label{app:mem0}

The scaled mem0 test uses separate user identifiers and one dated source
conversation per case. mem0 stored all 128 progressive facts and flattened 49.
By frame, the counts are affiliate role 9/16, class 6/16, equipment 1/16,
household 1/16, lodging 0/16, project 0/16, vehicle 16/16, and workplace 16/16.
It added no progressive wording or ``currently'' to 128 simple controls.
Date anchors appear in 109 progressive notes and 49 simple notes. In two additional
input sets, it removed ``for now'' from 91 of 128 notes and ``at the
moment'' from 104 of 128.

The same 512 inputs were also run with \texttt{google/gemini-3-flash-preview}
as mem0's writer, with everything else unchanged. Table~\ref{tab:mem0-writers}
compares the two writers. Gemini 3 Flash flattened 15 of 128 progressive statements
and added ``currently'' to 16 of 128 simple statements, all in the equipment
frame. A stratified hand-read of 36 Gemini 3 Flash notes agreed with every automatic
label.

\begin{table}[!htbp]
\centering
\footnotesize
\setlength{\tabcolsep}{2.2pt}
\begin{tabular}{@{}lrrrrrr@{}}
\toprule
 & \multicolumn{2}{c}{Progressive} & \multicolumn{2}{c}{``at the moment''} & \multicolumn{2}{c}{``for now''} \\
\cmidrule(lr){2-3}\cmidrule(lr){4-5}\cmidrule(l){6-7}
Frame & \mico{deepseek-color}DS & \mico{gemini-color}Gem & \mico{deepseek-color}DS & \mico{gemini-color}Gem & \mico{deepseek-color}DS & \mico{gemini-color}Gem \\
\midrule
Vehicle & 16 & 14 & 14 & 0 & 15 & 1 \\
Workplace & 16 & 1 & 16 & 1 & 11 & 4 \\
Equipment & 1 & 0 & 7 & 0 & 7 & 0 \\
Affiliate role & 9 & 0 & 16 & 0 & 13 & 4 \\
Class & 6 & 0 & 14 & 0 & 9 & 13 \\
Lodging & 0 & 0 & 16 & 0 & 13 & 0 \\
Household & 1 & 0 & 9 & 0 & 10 & 0 \\
Project & 0 & 0 & 12 & 0 & 13 & 0 \\
\midrule
All (of 128) & 49 & 15 & 104 & 1 & 91 & 22 \\
\bottomrule
\end{tabular}
\caption{Temporal cues lost in mem0 notes by writer model (DeepSeek V4 Flash,
Gemini 3 Flash), out of 16 inputs per frame. Progressive counts flattened
notes; the other columns count notes that lost the phrase.}
\label{tab:mem0-writers}
\end{table}

\section{Exploratory Reader Details}
\label{app:reader-details}

The dated-note test uses 96 source facts across eight situations. Each reader model receives
simple-present, progressive, and explicitly bounded notes at fresh and stale gaps, first
as a 0--100 persistence question and then as a field fill with
\textsc{unknown} available. This produces 1,152 requests per reader model. DeepSeek
V4 Flash ran on the official DeepSeek endpoint; GPT-5.6 Luna and GLM-5.2 ran
through OpenRouter with the quantization fixed but the serving host chosen by
the router. No request failed, and one unparsed DeepSeek field fill was
excluded. A hand-read found 60/60 parser decisions correct.
Table~\ref{tab:reader-ladder} gives the stale-note comparison for each
reader.

\begin{table}[!htbp]
\centering
\scriptsize
\setlength{\tabcolsep}{2.2pt}
\begin{tabular}{@{}lrrrrr@{}}
\toprule
Reader & Persistence diff. & Simple & Prog. & $b:c$ & $p$ \\
\midrule
\mico{deepseek-color}\deepseekname{DeepSeek, official} & -31.5 & 29 & 6 & 26:3 & $1.5{\times}10^{-5}$ \\
\mico{openai}\lunaname{Luna, OpenRouter} & -16.1 & 50 & 42 & 23:15 & .26 \\
\mico{zai}\glmname{GLM, OpenRouter} & -18.5 & 28 & 14 & 17:3 & .003 \\
\bottomrule
\end{tabular}
\caption{Stale progressive versus simple-present notes in the dated-note test
($n=96$ pairs per reader model). Persistence diff.\ is the progressive
minus the simple-present persistence estimate. Simple and Prog.\ count filled
fields; $b:c$ counts pairs filled only for the simple versus only for the
progressive note.}
\label{tab:reader-ladder}
\end{table}

Fresh simple-present notes receive mean persistence 94.8--97.4 and are used in 87--96
of 96 cases. Expired bounded notes receive means 0--2.4 and are refused in all
96 cases by each reader model. The stale simple-present form is itself used in only 28--50 of
96 cases, so the displayed date and \textsc{unknown} option drive much of the
abstention. Repeating the persistence question reproduces the stale
progressive-minus-simple-present contrast (-29.3, -15.6, and -18.7). DeepSeek's mean
absolute change between identical calls is 4.9 points on the official endpoint
but 15.7 points between the official and OpenRouter runs. On OpenRouter,
DeepSeek's field-fill contrast is null despite a similar persistence
contrast.

The minimal-edit test isolates the stored verb while holding the note format,
date, reader prompt, key fact, and response options fixed. We mechanically
restore progressive form in notes written from simple-present statements, or
flatten progressive form in notes where mem0 preserved it, then hand-check
every edit. Table~\ref{tab:reader-edits}
reports the paired results. The persistence $p$-values use two-sided Wilcoxon
signed-rank tests; the field-fill values use exact two-sided McNemar tests.

\begin{table}[!htbp]
\centering
\scriptsize
\setlength{\tabcolsep}{2.2pt}
\resizebox{\columnwidth}{!}{%
\begin{tabular}{@{}llrrrrr@{}}
\toprule
Edit & Reader & $n$ & $\Delta P$ ($p$) & Fills before$\rightarrow$after & $b:c$ & $p$ \\
\midrule
\multirow{3}{*}{Restore prog.}
& \mico{deepseek-color}\deepseekname{DeepSeek} & 37 & $-13.9$ (.00034) & $15\rightarrow3$ & 13:1 & .0018 \\
& \mico{openai}\lunaname{Luna} & 37 & $-6.1$ (.00045) & $18\rightarrow18$ & 6:6 & 1.0 \\
& \mico{zai}\glmname{GLM} & 37 & $-9.4$ ($1.3{\times}10^{-6}$) & $20\rightarrow15$ & 10:5 & .30 \\
\midrule
\multirow{3}{*}{Flatten prog.}
& \mico{deepseek-color}\deepseekname{DeepSeek} & 57 & $+37.3$ ($2.6{\times}10^{-8}$) & $1\rightarrow4$ & 1:4 & .375 \\
& \mico{openai}\lunaname{Luna} & 57 & $+18.2$ ($1.5{\times}10^{-6}$) & $16\rightarrow22$ & 7:13 & .263 \\
& \mico{zai}\glmname{GLM} & 57 & $+24.6$ ($7.9{\times}10^{-10}$) & $0\rightarrow2$ & 0:2 & .50 \\
\bottomrule
\end{tabular}
}
\caption{Minimal edits to mem0's stored notes. $\Delta P$ is the persistence
estimate after minus before. For field fill, $b:c$ counts items filled only
before versus only after the edit. Endpoints follow
Table~\ref{tab:reader-ladder}; tests are exploratory.}
\label{tab:reader-edits}
\end{table}

\subsection{Verification-tool task}
\label{app:tool}

The reader receives one dated note, a user request, and two tools: one
completes the request with the remembered value, and one asks the user to
confirm it first. The system prompt tells the reader to act if the stored value
can be relied on as still current and to ask for confirmation if it may no
longer be current. Only the first
tool call is scored, and a call counts as acting only if it uses the stored
value exactly. In each reader's main-test cell, all 30 notes written the same day
were acted on and all 30 notes that had explicitly expired were confirmed.

The main test crosses the 120 same-verb contents of \lapse{} with two or
three request wordings, for 291 pairs per reader. Lodging and household are
excluded because their edits change the verb. The analysis was written before
any target call. It uses a one-sided exact McNemar test per pair and a one-sided
sign test per content, each corrected across the three readers. The prediction
counted as confirmed only if GLM-5.2 at eight months passed both. GLM-5.2 and DeepSeek V4.1 Flash were
tested at note ages set from the calibration runs below. The eight-month GLM
row of Table~\ref{tab:tool-age} was itself a planned test on 69 pairs; it was not significant after correction ($p=.051$), and the main
test is its larger repetition. The age choice used only the rate of acting on
simple-present notes, but the same runs produced the form contrast that
motivated the repetition. The 69 facts of those runs are among the 120 used
here, with new request wordings. Gemini 3 Flash was
admitted by a sweep on 30 separate contents, which placed it at one month.
Three candidates could not be tested: GPT-5.6 Luna confirms at every age,
Mistral Small 3.2 confirmed only 12 of 30 expired notes, and Qwen3.8-Max
rejects a required tool call.

After the main test we added three readers. Each used the same items, gates,
age rule, and analysis, and each is analyzed on its own. Claude Haiku 5.5 and
Gemma 4 31B passed admission but fell outside the age rule. Haiku confirmed
every simple-present note at one, three, and eight months, and Gemma acted on
every one. Kimi K2.6, on the Moonshot AI endpoint, was placed at three months,
where it acted on 18 of 30 simple-present notes, and it passed its cell gate. DeepSeek V4.1 Flash ran on the official endpoint;
it is a later model than the DeepSeek V4 Flash used elsewhere in the paper.
Table~\ref{tab:tool-powered} gives the result.

\begin{table}[!htbp]
\centering
\scriptsize
\setlength{\tabcolsep}{1.6pt}
\resizebox{\columnwidth}{!}{%
\begin{tabular}{@{}lrrrrrrl@{}}
\toprule
Reader & Months & \shortstack{Simple\\acted on} & \shortstack{Progressive\\acted on} & \shortstack{Simple only :\\prog.\ only} & \shortstack{Contents\\$+$ : $-$} & \shortstack{Difference\\(95\% CI)} & \shortstack{Corrected\\$p$} \\
\midrule
\mico{zai}GLM & 8 & 187/291 & 116/291 & 91:19 & 54:9 & $+25$ (18, 32) & $2.8{\times}10^{-12}$ \\
Gemini 3 Flash & 1 & 135/291 & 114/291 & 23:2 & 19:0 & $+7$ (4, 11) & $1.9{\times}10^{-5}$ \\
\mico{deepseek-color}DeepSeek 4.1 & 1 & 155/291 & 144/291 & 30:19 & 23:13 & $+4$ ($-1$, 8) & $.076$ \\
Kimi K2.6$^\dagger$ & 3 & 169/291 & 191/291 & 34:56 & 23:36 & $-8$ ($-15$, $-0.4$) & $.026^\dagger$ \\
\bottomrule
\end{tabular}}
\caption{Verification-tool main test (291 pairs, 120 contents per reader).
``Contents $+$ : $-$'' counts contents whose pairs lean toward acting on the
simple-present or the progressive note. The difference is in percentage
points, with a bootstrap interval over contents. $p$ is the one-sided exact
McNemar test, Holm-corrected across the three readers. GLM ran on Baidu (fp8)
through OpenRouter, Gemini 3 Flash on Google AI Studio, DeepSeek on its official
endpoint. Three GLM pairs with a truncated value are excluded.
$^\dagger$Added after the main test, with a two-sided exact $p$, uncorrected.
Three unparsed Kimi pairs are excluded.}
\label{tab:tool-powered}
\end{table}

GLM's per-pair splits by kind of fact are 27:1 for affiliate role, 31:0 for
class, and 12:4 for workplace. They are 12:7 for project and 6:1 for vehicle.
Equipment reverses at 3:6, and both of its forms are rarely acted on. Every
request wording leans the same way. Gemini 3 Flash never acts on a class note in
either form. DeepSeek's project notes go the other way (4:10). Kimi K2.6 leans
toward the progressive note for projects (5:20) and affiliate roles (6:16). It
leans toward the simple note for workplaces (8:3) and acts on every class
note in both forms.

The mem0 test uses notes written by installed mem0 with DeepSeek V4 Flash as
the writer (Section~\ref{sec:mem0}). In the preserved subset, mem0 kept the
progressive and we flatten the verb by hand (34 pairs). The simple-source
subset covers the contents where mem0 flattened the progressive statement; we
take the note mem0 wrote from the matched simple-present statement and restore
the progressive (37 pairs). The six
tests are corrected as one family (Table~\ref{tab:tool-mem0}). For GLM, the
verb matters on stored lines without a date, which split 14:1 in the
simple-source subset. Lines with a date split 1:1 in both subsets. Gemini 3 Flash's
four reversed pairs in the preserved subset are all equipment or project notes.

\begin{table}[!htbp]
\centering
\scriptsize
\setlength{\tabcolsep}{2pt}
\resizebox{\columnwidth}{!}{%
\begin{tabular}{@{}llrrrr@{}}
\toprule
Reader & Subset & \shortstack{Simple\\acted on} & \shortstack{Progressive\\acted on} & \shortstack{Simple only :\\prog.\ only} & \shortstack{Corrected\\$p$} \\
\midrule
\mico{zai}GLM, 8 mo & preserved & 9/34 & 7/34 & 4:2 & 1 \\
 & simple source & 30/37 & 17/37 & 15:2 & .007 \\
\addlinespace
Gemini 3 Flash, 1 mo & preserved & 10/34 & 14/34 & 0:4 & 1 \\
 & simple source & 19/37 & 14/37 & 5:0 & .16 \\
\addlinespace
\mico{deepseek-color}DeepSeek 4.1, 1 mo & preserved & 16/34 & 11/34 & 6:1 & .25 \\
 & simple source & 14/37 & 13/37 & 4:3 & 1 \\
\addlinespace
Kimi K2.6$^\dagger$, 3 mo & preserved & 21/34 & 19/34 & 3:1 & .63 \\
 & simple source & 28/37 & 26/37 & 5:3 & .63 \\
\bottomrule
\end{tabular}}
\caption{Verification-tool test on mem0's stored notes. ``Simple'' is the
flattened or simple-source note. $p$ is the one-sided exact McNemar test,
Holm-corrected across the six tests. One DeepSeek preserved pair is
unparsed. $^\dagger$Added after the main test; Holm-corrected across its own
two tests.}
\label{tab:tool-mem0}
\end{table}

Before the main test, the same task ran on 69 pairs whose notes came from
Graphiti, at one, three, and eight months (Table~\ref{tab:tool-age}). Its
eight-month GLM cell was a planned test and did not pass the
correction (raw $p=.017$). These runs set the note ages of GLM and DeepSeek
in the main test, and 69 of its 120 contents appear in them with other request
wordings.

\begin{table}[!htbp]
\centering
\scriptsize
\setlength{\tabcolsep}{1.8pt}
\begin{tabular}{@{}llrrrrl@{}}
\toprule
Reader & Months & \shortstack{Simple\\acted on} & \shortstack{Progressive\\acted on} & \shortstack{Simple only :\\prog.\ only} & $p$ & Note \\
\midrule
\mico{deepseek-color}DeepSeek 4.1 & 1 & 33/69 & 28/69 & 5:0 & .062 &  \\
 & 3 & 20/69 & 14/69 & 9:3 & .146 &  \\
 & 8 & 5/69 & 2/69 & 5:2 & .453 & bound. \\
\addlinespace
\mico{zai}GLM & 1 & 61/69 & 66/69 & 1:4 & .375 & bound. \\
 & 3 & 59/69 & 50/69 & 14:3 & .013 &  \\
 & 8 & 38/69 & 28/69 & 17:5 & .017 &  \\
\addlinespace
\mico{openai}Luna & 1 & 1/69 & 0/69 & 1:0 & 1.000 & bound. \\
 & 3 & 1/69 & 0/69 & 1:0 & 1.000 & bound. \\
 & 8 & 0/69 & 0/69 & 0:0 & 1.000 & bound. \\
\bottomrule
\end{tabular}
\caption{Verification-tool calibration runs by note age (69 same-verb pairs per cell). ``Acted
on'' counts first tool calls that complete the request with the remembered
value instead of asking for confirmation. $p$ is an unadjusted two-sided exact
McNemar test. ``Bound.'' marks a floor or ceiling cell (an arm under 10\% or
over 90\%). DeepSeek 4.1 is DeepSeek V4.1 Flash.}
\label{tab:tool-age}
\end{table}

\subsection{Free-text note and structured record}
\label{app:record}

This test was added after the main experiments and is exploratory. Its analysis
was written before any reader call. DeepSeek V4 Flash wrote notes for the 120
contents of Appendix~\ref{app:tool} in progressive and simple-present form,
from the memory-write session of Appendix~\ref{app:materials}. Each  
session received the baseline memory-write query, and separately the same query
with this instruction appended: \emph{Write each note as a record with exactly
these four fields, one per line: said: the user's own words, quoted exactly;
fact: the fact in a short normalized form; form: the grammatical form of what
the user said (simple present, present progressive, past, future, or other);
observed: the session date.} The reader saw the line or record that names the
fact.

In free text, the writer rendered 99 of the 120 progressive sources as
simple-present states. In records, the \texttt{said} field contained the user's
sentence for all 240 notes. The \texttt{form} field named the progressive for
112 of the 120 progressive sources and the simple present for all 120
simple-present sources. The \texttt{fact} field kept the progressive in 24
notes.

Each reader saw four versions of every note in the verification-tool task: the
free-text line, the full record, the record without the \texttt{said} field,
and the user's sentence alone (``user said on [date]: \ldots''). Gemini 3 Flash
read one-month notes and GLM-5.2 eight-month notes, the ages of
Table~\ref{tab:tool-powered}, on the same endpoints. Both passed a gate of 30
same-day and 30 expired notes written as records. Two tests were fixed before
the reader calls. The first asks whether the form contrast is larger for the
record than for the free-text note (one-sided sign test over contents). The
second asks whether, on progressive sources, the reader's decision matches its
decision on the user's sentence more often with the record than with free text
(one-sided exact McNemar test). We ran Gemini 3 Flash first and decided to run
GLM-5.2 after seeing its result.

The first test passed for both readers. For Gemini 3 Flash, 35 contents leaned
toward a larger contrast for the record and 6 the other way
($p=2.4{\times}10^{-6}$). For GLM-5.2 the split was 64 to 13
($p=1.5{\times}10^{-9}$).

The second test passed for GLM-5.2. Its decisions on records matched its
decisions on the user's sentence 218 times, against 176 for free text
($p=3.2{\times}10^{-5}$). Asked about the user's sentence a second time, GLM-5.2
repeated its first decision 221 times, so the record matched about as often as
a repeat of the same input. The second test failed for Gemini 3 Flash (206
matches against 212). Each version set its own overall rate of acting
(Table~\ref{tab:record}), which limits how often any version can match the
user's sentence. On simple-present sources, the record matched the user's sentence less often than free text for both readers.

\begin{table}[!htbp]
\centering
\scriptsize
\setlength{\tabcolsep}{2pt}
\begin{tabular}{@{}llrrrr@{}}
\toprule
Reader & Note shown & \shortstack{Simple\\acted on} & \shortstack{Progressive\\acted on} & \shortstack{Simple only :\\prog.\ only} & $p$ \\
\midrule
\mico{zai}GLM-5.2 & Free text & 160/278 & 156/278 & 23:19 & $.32$ \\
 & Record & 177/287 & 82/287 & 113:18 & $2.6{\times}10^{-18}$ \\
 & Record without words & 157/285 & 133/285 & 49:25 & $.0035$ \\
 & User's sentence & 137/279 & 77/279 & 75:15 & $4.6{\times}10^{-11}$ \\
Gemini 3 Flash & Free text & 129/291 & 128/291 & 11:10 & $.50$ \\
 & Record & 180/291 & 138/291 & 45:3 & $6.6{\times}10^{-11}$ \\
 & Record without words & 161/291 & 153/291 & 17:9 & $.084$ \\
 & User's sentence & 94/291 & 53/291 & 47:6 & $2.9{\times}10^{-9}$ \\
\bottomrule
\end{tabular}
\caption{The verification-tool task on notes written by DeepSeek V4 Flash as
free text or as records. GLM-5.2 read eight-month notes and Gemini 3 Flash
one-month notes. Pairs with an unparsed response are excluded. $p$ is a
one-sided exact McNemar test, uncorrected.}
\label{tab:record}
\end{table}

\FloatBarrier
\section{Pipeline-to-Reader Details}
\label{app:chain-details}

Each pipeline receives 288 isolated inputs: 96 key facts (witnesses) in
progressive, simple-present, and explicitly bounded form. The writer clock is fixed to
2026-01-06, and readers evaluate the stored witness-bearing line on
2026-09-08. A line is labeled preserved when it contains a progressive verb or
an explicit ongoing marker such as ``currently''; nominal and simple-present
rewrites are labeled flattened. Missing witness lines are dropped and counted.
Table~\ref{tab:chain-writers} gives the writer outcomes, and
Table~\ref{tab:channel-survival} how often each reader used the stored facts.

\begin{table}[!htbp]
\centering
\scriptsize
\setlength{\tabcolsep}{2.5pt}
\begin{tabular}{@{}lrrrr@{}}
\toprule
Pipeline/form & Stored & Preserved & Flattened & Dropped \\
\midrule
mem0/progressive & 94 & 57 & 37 & 2 \\
mem0/simple & 96 & 0 & 96 & 0 \\
mem0/bounded & 94 & 76 & 18 & 2 \\
Graphiti/progressive & 95 & 48 & 47 & 1 \\
Graphiti/simple & 96 & 0 & 96 & 0 \\
Graphiti/bounded & 96 & 54 & 42 & 0 \\
Letta/progressive & 90 & 28 & 62 & 6 \\
Letta/simple & 93 & 1 & 92 & 3 \\
Letta/bounded & 95 & 18 & 77 & 1 \\
\bottomrule
\end{tabular}
\caption{Writer outcomes in the pipeline-to-reader experiments. Bounded lines
can be labeled flattened after losing progressive morphology while still
retaining their explicit end date.}
\label{tab:chain-writers}
\end{table}

\begin{table}[!htbp]
\centering
\scriptsize
\setlength{\tabcolsep}{2.2pt}
\resizebox{\columnwidth}{!}{%
\begin{tabular}{@{}lllrrrrrr@{}}
\toprule
Reader & Pipeline & Cue & $n$ & Prog. used & Simple used & Simple only & Prog. only & Raw $p$ \\
\midrule
\multirow{5}{*}{\mico{deepseek-color}\deepseekname{DeepSeek}}
& mem0 & verb & 57 & 1 & 12 & 12 & 1 & .003 \\
& Graphiti & verb & 48 & 4 & 19 & 18 & 3 & .0015 \\
& Graphiti & none & 47 & 18 & 25 & 13 & 6 & .17 \\
& Letta & ``Currently'' & 28 & 0 & 6 & 6 & 0 & .031 \\
& Letta & none & 62 & 16 & 22 & 13 & 7 & .26 \\
\midrule
\multirow{5}{*}{\mico{openai}\lunaname{Luna}}
& mem0 & verb & 57 & 16 & 21 & 12 & 7 & .36 \\
& Graphiti & verb & 48 & 16 & 29 & 15 & 2 & .0024 \\
& Graphiti & none & 47 & 22 & 28 & 9 & 3 & .15 \\
& Letta & ``Currently'' & 28 & 10 & 12 & 7 & 4 & .55 \\
& Letta & none & 62 & 40 & 42 & 12 & 9 & .66 \\
\midrule
\multirow{5}{*}{\mico{zai}\glmname{GLM}}
& mem0 & verb & 57 & 0 & 10 & 10 & 0 & .002 \\
& Graphiti & verb & 48 & 1 & 13 & 13 & 1 & .0018 \\
& Graphiti & none & 47 & 22 & 28 & 12 & 6 & .24 \\
& Letta & ``Currently'' & 28 & 0 & 8 & 8 & 0 & .0078 \\
& Letta & none & 62 & 23 & 32 & 16 & 7 & .093 \\
\bottomrule
\end{tabular}%
}
\caption{Reader use of each pipeline's stored facts. ``Simple only'' and
``Prog. only'' are discordant matched pairs. Values are exact McNemar tests.
Under Holm correction across these 15 tests, five remain below .05: the mem0
verb rows for DeepSeek and GLM and the Graphiti verb rows for all three
readers.}
\label{tab:channel-survival}
\end{table}

mem0 2.0.19 and Graphiti 0.30.1 use DeepSeek on the official endpoint; Letta
0.16.8 uses DeepSeek through OpenRouter. Reader endpoints match Appendix~\ref{app:reader-details}. The
mem0 test contains 1,064 rows per reader with no unparsed outputs; the Graphiti
and Letta test contains 1,704 rows per reader with four unparsed outputs in
total. Independent spot-checks found 45/45 field fills correctly parsed in each
test.

The mem0 prediction that flattened notes would produce no reader-model contrast
fails in two readers because mem0 moves the cue into a date stamp. For Graphiti,
the predicted null on its 47 cue-absent facts holds in all three readers,
as does the predicted contrast on its 48 verb-preserved facts. Letta's pooled
null prediction fails in two readers because 28 lines retain a ``Currently''
prefix; the 62 bare-label lines are null in all three. These nulls do not prove
equivalence: their raw counts lean in the predicted direction, and only 40 of
47 Graphiti pairs and 11 of 61 comparable Letta pairs are string-identical.
Graphiti sets \texttt{valid\_at} to the write time on all 287 witness edges and
never sets \texttt{invalid\_at} for a progressive input. It does extract an
explicit endpoint from 91 of 96 bounded inputs, contrary to our prediction.

\FloatBarrier
\section{Benchmark Screen}
\label{app:ecology}

The LongMemEval input is deduplicated to 1,744 user-turn segments; LoCoMo
contributes 2,486 turns. A high-recall marker search followed by contextual
inspection yields 80 unique utterances containing progressive or perfect
marking. The search was designed to find examples and its recall is unknown,
so this count is not a rate. We read the benchmark questions about these
utterances and found none that can be answered only by noticing that the
statement was in temporary form: when a fact changes, a later turn says so, or
the most recent mention answers the question. This is the authors' reading of
a set selected by the search, not a blind count.

To check that the marked forms occur outside curated benchmarks, we also
ran the same marker search over 809,281 English user turns of WildChat-1M
\citep{zhao2024wildchat}. A hit is a regular-expression match and nothing
more. We read all 339 progressive and perfect-progressive hits under a fixed
three-question rule (the user's own first-person statement; a current state
with a plausible end; a life circumstance rather than the tool stack of the
current question). This yields 57 distinct life-circumstance facts in
temporary form, 14 of which match a \lapse{} frame almost exactly (``from
yesterday I'm staying at my parent's place''; ``I have been taking fluoxetine
for approximately 21 months''), and about 200 tool-in-use statements. The
bounded and adverbial marker classes had near-zero precision and are
excluded. This is an occurrence check, not a rate: recall is unknown, the
read was performed by the authors rather than blind annotators, and
WildChat users state such facts only when a task requires them. The rule and
the screening code are released; the excerpts are not redistributed.

\subsection{Writer test on screened utterances}

We also passed 104 screened life-circumstance statements through the three
confirmatory writers. The set contains WildChat utterances and dialogue from
LoCoMo and LongMemEval. All outputs labeled dropped or flat were hand-read.
Table~\ref{tab:real-utterances} gives the results.

\begin{table}[!htbp]
\centering
\small
\setlength{\tabcolsep}{3pt}
\begin{tabular}{@{}lrrr@{}}
\toprule
Source form & \mico{deepseek-color}DeepSeek & \mico{openai}Luna & \mico{zai}GLM \\
\midrule
Bare present progressive & 9/27 & 6/28 & 5/29 \\
Bare perfect progressive & 13/27 & 14/27 & 12/26 \\
Duration-bearing & 5/45 & 4/45 & 8/44 \\
\bottomrule
\end{tabular}
\caption{Loss of temporal qualification in stored real-utterance facts.
Denominators include only stored facts. The regex screen is not a prevalence
sample, and labels were read by one analyst.}
\label{tab:real-utterances}
\end{table}

\FloatBarrier
\section{Reproducibility and Release}
\label{app:repro}

The confirmatory protocol was frozen internally on 2026-08-12, before scored
outcomes were opened. It was not deposited in a third-party registry, so we
describe it as fixed before data collection, not as preregistered. The freeze  
is commit \texttt{a60dcc5b5c0f555b}\allowbreak
\texttt{05c3d766e8f50cf3}\allowbreak
\texttt{24250c3a}; the frozen specification has SHA-256
\texttt{8521448b3f012eba}\allowbreak
\texttt{c0f131bd2c200e7f}\allowbreak
\texttt{14041fb4816e7307}\allowbreak
\texttt{a137ec8b86053deb}.
The released specification also includes later dated amendments. The
estimation rows and exploratory experiments they added are not part of the
confirmatory tests.

The release (\lapseurl) includes the deterministic stimulus builder, reference and run
manifests, raw response traces, and cascade and judge outputs. It also
includes the blinded gold labels and adjudication rules, the two analysis
programs with their expected byte outputs, the mem0 traces, and a
claim-to-artifact hash ledger. The generator is
v2.0.1 (source SHA-16 \texttt{58fb635d3b996e71}); the reference stimulus file
has SHA-16 \texttt{87561af0713a9fff}. Analysis programs are rerun before any
figure build, and the build aborts unless their textual outputs reproduce the
committed SHA-16 values exactly. Every plotted count is then derived from
hash-verified artifacts and serialized to one figure-data JSON file.

All names, companies, institutions, projects, and residences in \lapse{} are
invented. A contamination screen found no strong web collisions after rotating
three weak names. Model outputs can still contain provider-generated text, so
the public release should preserve applicable provider terms and redact any
request metadata not needed for reproduction. The benchmark evaluates memory
systems, not users, and contains no personal conversations.

\end{document}